\documentclass{article}

\usepackage{microtype}
\usepackage{graphicx}
\usepackage{subcaption}
\usepackage{booktabs} 

\usepackage{tabularx}
\usepackage{multirow}

\usepackage{threeparttable}

\usepackage{array, listings}

\usepackage{hyperref}

\usepackage[preprint]{icml2026}

\usepackage{amsmath}
\usepackage{amssymb}
\usepackage{mathtools}
\usepackage{amsthm}

\usepackage{algorithm}
\usepackage{algpseudocode}

\usepackage[capitalize,noabbrev]{cleveref}

\theoremstyle{plain}

\theoremstyle{definition}

\theoremstyle{remark}

\usepackage[textsize=tiny]{todonotes}

\icmltitlerunning{STaSBO: LLM-Guided Strategy Planning and Configuration Selection for Bayesian Optimization}

\begin{document}

\twocolumn[
  \icmltitle{STaSBO: LLM-Guided Strategy Planning and \\Configuration Selection for Bayesian Optimization}

  \icmlsetsymbol{equal}{*}

  \begin{icmlauthorlist}
    \icmlauthor{Youngseok Yoon}{ucsb}
    \icmlauthor{Peng Li}{ucsb}
  \end{icmlauthorlist}

  \icmlaffiliation{ucsb}{Department of Electrical and Computer Engineering, University of California, Santa Barbara, California, United States}

  \icmlcorrespondingauthor{Peng Li}{lip@ucsb.edu}
  
  \icmlkeywords{Bayesian optimization, LLM, Agentic-AI}

  \vskip 0.3in
]



\printAffiliationsAndNotice{}  

\begin{abstract}
  Bayesian optimization (BO) enables sample-efficient search for expensive black-box problems, but its performance depends strongly on surrogate--acquisition configurations whose suitability varies across problems and optimization states. We take a significant step toward \emph{LLM-driven strategic management of BO} with \textbf{STaSBO}, a task-agnostic meta-controller that plans search behavior and jointly selects surrogate--acquisition configurations, while numerical BO retains responsibility for probabilistic modeling and candidate generation. STaSBO comprises two coordinated stages: \emph{strategy planning} and \emph{strategy-guided configuration selection}. A structured memory system connects prior plans, configuration selections, and candidate behaviors to optimization states, allowing the controller to revise its trajectory summary and behavioral plans during planning. To navigate the large joint configuration space, the selection stage shortlists configurations using surrogate fit evidence. Prospective BO-action diagnostics then expose the candidate behaviors, supporting final joint selection against the behavioral plans. Across 12 synthetic mathematical and 10 scientific and engineering application tasks, STaSBO achieves the best average rank against 11 baselines: 2.17 and 2.60, compared with 4.08 and 4.10 for the strongest baseline in each suite, respectively.
\end{abstract}

\section{Introduction}
\label{section:introduction}

Scientific and engineering design often requires costly black-box evaluations under limited budgets~\citep{shields2021bayesian, macleod2022self}. Bayesian optimization (BO) uses a probabilistic surrogate and acquisition function to guide sample-efficient search~\citep{shahriari2015taking, snoek2012practical, frazier2018tutorial}, with demonstrated practical benefits~\citep{burger2020mobile, shields2021bayesian}. Yet its effectiveness depends critically on the surrogate--acquisition \emph{configuration}~\citep{liang2021benchmarking}. Our sensitivity analysis shows that fixed configurations can transfer poorly across problems and that the benefit of switching configurations during optimization depends on the problem, switching time, and destination configuration. Configuration selection is thus an online decision that shapes the search trajectory, not merely an initialization choice.

\begin{figure*}[h]
    \centering
    \includegraphics[width=\textwidth]{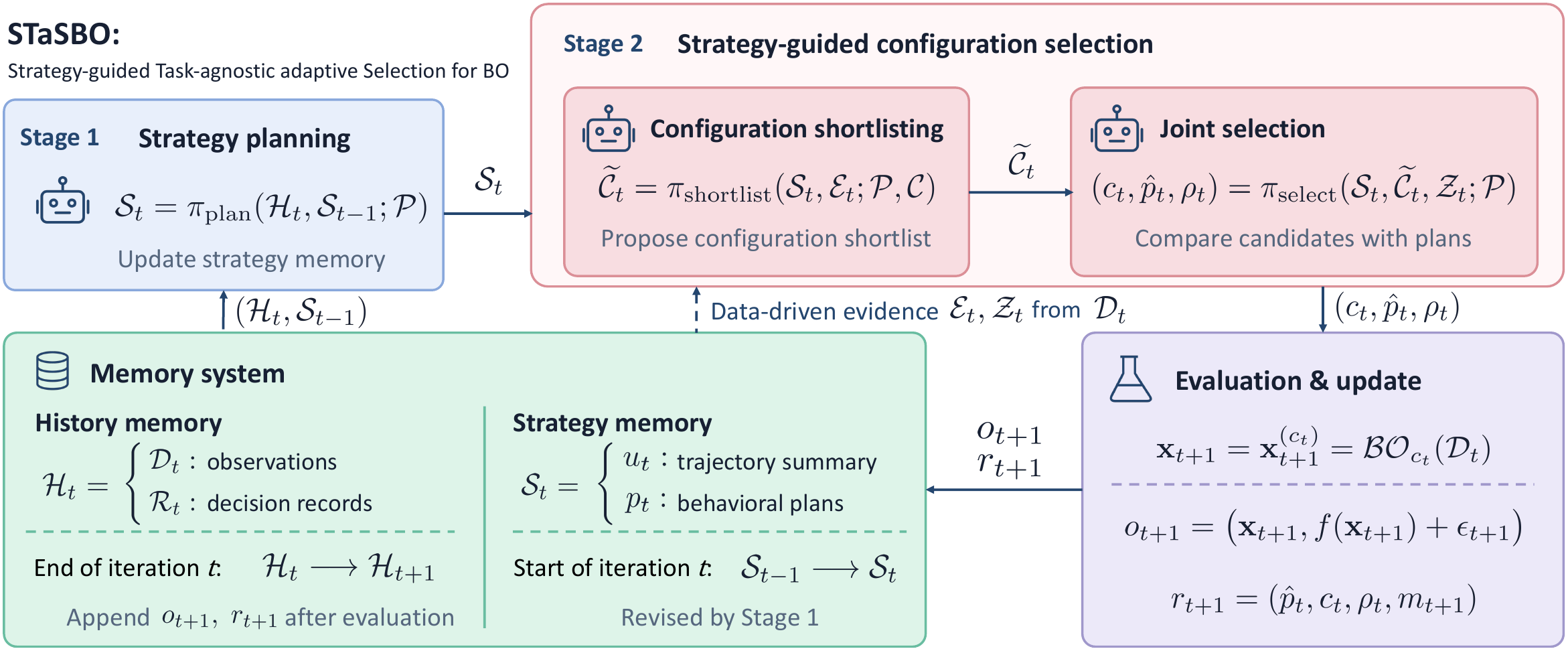}
    \caption{Overview of STaSBO. Stage 1 (strategy planning) updates trajectory summary and behavioral plans from historical and previous strategy memory. Stage 2 (strategy-guided configuration selection) shortlists configuration sets using surrogate fit evidence and updated strategy, then compares candidates from shortlisted sets using prospective BO-action diagnostics. Only the selected candidate is evaluated, and its observation and decision record inform the next iteration.}
    \label{fig:llm_bo_framework}
\end{figure*}

Existing methods address parts of this challenge. Meta-learning and prior-based BO establish surrogates or acquisition policies using related tasks or sampled function priors~\citep{volpp2019meta, wistuba2021few, muller2023pfns4bo, wang2024pre}. Acquisition portfolios switch among acquisition functions as observations update reward scores, but follow predefined selection rules~\citep{hoffman2011portfolio, vasconcelos2019no}. Recent LLM approaches directly propose or select evaluation points using problem descriptions and numerical search histories~\citep{yang2024large, liu2024large, cisse2025language, chang2025llinbo, yang2025reasoning}. Others use LLMs to adapt individual BO components: CAKE evolves kernels under a prescribed acquisition rule~\citep{suwandi2026adaptive}, whereas LMABO selects acquisitions within a prescribed surrogate family~\citep{ngo2026adaptive}. These methods demonstrate useful forms of adaptation using LLMs, but direct point generation does not exploit BO's numerical candidate-generation machinery, while adapting only one component leaves search behavior restricted to the fixed other components.

We take a significant step toward \emph{LLM-driven strategic management of BO}. Rather than proposing evaluation points directly, the LLM serves as a high-level, task-agnostic meta-controller that interprets search progress, plans subsequent search behaviors, and jointly selects a surrogate--acquisition configuration; numerical BO retains responsibility for probabilistic modeling and candidate generation. This combines contextual reasoning of LLMs with numerical optimization of BO. Joint control accounts for the fact that an acquisition's behavior depends on its fitted surrogate. Moreover, static profile descriptions alone do not reveal where a configuration will query under the current observations. The central challenge remains: determining what the search should accomplish next and which configuration can realize that intention.

We propose \textbf{STaSBO} (\emph{Strategy-guided Task-agnostic adaptive Selection for Bayesian Optimization}), a meta-controller addressing this challenge through two coordinated stages: \emph{strategy planning} and \emph{strategy-guided configuration selection} (Figure~\ref{fig:llm_bo_framework}). Planning makes search intentions explicit before configuration selection by examining how prior plans led to selected configurations, how those configurations shaped candidate behavior, and what progress followed. A structured \emph{memory system} maintains observations and decision records alongside a revisable trajectory summary and behavioral plans. This feedback allows the controller to distinguish whether an intended behavior was realized from whether the objective improved, and to retain or revise its strategy accordingly.

The second stage grounds those plans in current observations through \emph{surrogate fit evidence} and \emph{prospective BO-action diagnostics}. Numerical evidence from fitted surrogate models complements static profile specifications during configuration shortlisting, where the LLM proposes surrogate profiles together with acquisition families and parameter values. Numerical BO then generates prospective candidates for the shortlisted configurations without evaluating the objective. Diagnostics, including distances to the incumbent and previously evaluated points, expose each configuration's data-dependent candidate behavior. The LLM compares these prospective actions against the behavioral plans and jointly selects a configuration; only its candidate is evaluated. Together, the two stages connect strategy informed by past decisions to the actual search actions available now, with memory coordinating this process across iterations.

\begin{table*}[h]
    \centering
    \caption{Cross-problem (function $\times$ dim) transfer of oracle configurations. \textbf{Oracle configurations are problem dependent and do not transfer well across target problems.} For each source problem, ``Oracle Config.'' denotes the best-performing configuration among 30 fixed configurations. Entries report the rank of ``Oracle Config.'' by final simple regret on each target problem; lower is better.}
    \label{tab:configuration_transfer}
    
    \small
    \setlength{\tabcolsep}{4pt}
    \renewcommand{\arraystretch}{1.12}
    \begin{tabular*}{\linewidth}{@{\extracolsep{\fill}}lcc*{6}{c}@{}}
        \toprule
        \multirow{3}{*}{\shortstack{Source\\Function}} &
        \multirow{3}{*}{\shortstack{Source\\Dim}} &
        \multirow{3}{*}{\shortstack{Oracle\\config.}} &
        \multicolumn{2}{c}{Ackley} & \multicolumn{2}{c}{Levy} & \multicolumn{2}{c}{Rastrigin} \\
        \cmidrule(lr){4-5}\cmidrule(lr){6-7}\cmidrule(lr){8-9}
        & & & 5D & 15D & 5D & 15D & 5D & 15D \\
        \midrule
        \multirow{2}{*}{Ackley} & 5D  & S1--A5 & \textbf{1} & 26          & 22          & 25          & 24          & 27          \\
                                & 15D & S5--A3 & 15         & \textbf{1}  & 12          & 4           & 14          & 3           \\
        \addlinespace[1pt]
        \multirow{2}{*}{Levy}   & 5D  & S1--A4 & 27         & 10          & \textbf{1}  & 9           & 11          & 6           \\
                                & 15D & S5--A1 & 2          & 7           & 10          & \textbf{1}  & 5           & 2           \\
        \addlinespace[1pt]
        \multirow{2}{*}{Rastrigin} & 5D  & S5--A5 & 4          & 3           & 5           & 12          & \textbf{1}  & 4          \\
                                   & 15D & S2--A1 & 11         & 16          & 6           & 13          & 3           & \textbf{1}  \\
        \bottomrule
    \end{tabular*}
    
    \vspace{4pt}
    \begin{minipage}{\linewidth}
        \footnotesize\raggedright
        \textit{Surrogates:}
        S1, Mat\'ern-$5/2$ ARD;
        S2, Mat\'ern-$5/2$ ISO;
        S3, Mat\'ern-$3/2$ ARD;
        S4, RBF ARD;
        S5, RBF ISO.
        \par\smallskip
        \textit{Acquisitions:}
        A1--A3, EI ($\xi_{\mathrm{frac}}=0.00,0.05,0.20$, respectively);
        A4--A6, UCB ($\kappa=0.5,3.0,6.0$, respectively).
    \end{minipage}
    \vspace{0pt}
\end{table*}

Our contributions are summarized as follows:
\begin{itemize}
    \item We formulate task-agnostic, adaptive joint surrogate--acquisition selection as sequential LLM meta-control and develop a two-stage controller that separates strategy planning from configuration selection while retaining numerical BO for probabilistic modeling and candidate generation. The controller makes search intent explicit before jointly choosing surrogate profiles, acquisition families, and acquisition parameters, allowing configuration choices to adapt to the evolving optimization state.

    \item We introduce a decision-aware memory system connecting prior plans and configuration selections to search progress, together with surrogate fit evidence and prospective BO-action diagnostics that ground plan-conditioned shortlisting and final joint selection in current observations and candidate behavior. The feedback distinguishes whether intended search behavior was realized from whether the objective improved, providing complementary signals for revising subsequent strategy.

    \item We evaluate STaSBO against 11 baselines on 12 synthetic mathematical tasks and 10 scientific and engineering application tasks under a shared budget of 50 objective evaluations. STaSBO achieves the best average task rank in both suites: 2.17 on synthetic tasks and 2.60 on application tasks, compared with 4.08 and 4.10 for the best baseline in each suite, respectively. Component ablations further examine the contributions of joint configuration control, strategy planning, and numerical evidence.
\end{itemize}

\section{Background and motivation}
\label{section:background}

\subsection{Bayesian Optimization Preliminaries}

We consider an expensive black-box objective $f:\mathcal{X}\rightarrow\mathbb{R}$ over a $d$-dimensional design space $\mathcal{X}$, with noisy evaluations $y=f(\mathbf{x})+\epsilon$. Our goal is to find: $\mathbf{x}^{*}=\arg\max_{\mathbf{x}\in\mathcal{X}} f(\mathbf{x})$, starting from initial observations $\mathcal{D}_0 = \{(x_i^{init}, y_i^{init})\}_{i=1}^{n_0}$ and allowing $T$ additional evaluations. At iteration $t$, Bayesian optimization (BO) fits a probabilistic surrogate to $\mathcal{D}_t = \mathcal{D}_0 \cup \{(x_s, y_s) \}_{s=1}^{t}$ and selects the next evaluation by optimizing an acquisition function~\citep{shahriari2015taking}. A Gaussian process (GP) surrogate provides a predictive mean $\mu_t(\mathbf{x})$ and posterior standard deviation $\sigma_t(\mathbf{x})$, allowing the acquisition function to balance predicted performance and uncertainty~\citep{frazier2018tutorial}.

A \emph{surrogate profile} specifies the GP kernel, automatic relevance determination (ARD) setting, and associated hyperparameters~\citep{williams2006gaussian}; an \emph{acquisition profile} specifies the acquisition family and its hyperparameters. Together, they define a \emph{BO configuration}, whose candidate behavior depends on both components. Conventional BO typically fixes this configuration during optimization~\citep{snoek2012practical}.

\subsection{Sensitivity to Configuration}
\label{section:background:subsection:sensitivity}

We retrospectively evaluate 30 fixed surrogate--acquisition configurations on Ackley, Levy, and Rastrigin in 5D and 15D. Table~\ref{tab:configuration_transfer} identifies a distinct \emph{oracle configuration} for each of the six problems, where the oracle minimizes final simple regret averaged over seeds within this configuration pool. It is a diagnostic reference selected after completing all fixed-configuration runs, rather than an available online selection rule. Transfer performance varies substantially across functions and dimensionalities: the Ackley-5D oracle ranks 26th on Ackley-15D and between 22nd and 27th across the five other settings. Source-problem optimality therefore provides no assurance of comparable performance on target problems. Detailed setting and analysis are provided in Appendix~\ref{app:sensitivity_analysis}.

Figure~\ref{fig:configuration_sensitivity} examines whether a mid-run configuration switch can improve the subsequent search. For each problem, we use the configuration ranked tenth by fixed-run final mean regret as the reference (black trajectory) and branch its trajectory to each of the 29 alternatives after 15 or 30 iterations. Each branch inherits the reference observations and retains its new configuration through the common budget of 45 BO iterations. On Levy-15D, 20 of the 29 alternatives switched after 30 iterations achieve lower final mean simple regret than continuing the reference. On Rastrigin-5D, 13 alternatives improve on the reference when switched after 15 iterations, compared with eight after 30 iterations, although both switch groups have higher branch-averaged final regret than the reference. Switching can thus help, but its effectiveness depends on which configuration is chosen and when the change is made. Please refer to Appendix~\ref{app:sensitivity_analysis} for details.

Together, these observations motivate treating BO configuration adaptation as a sequential online decision process informed by the evolving optimization history. A configuration that succeeds on one problem may perform poorly on another, while changing configurations during a run can improve or degrade subsequent performance depending on the switch time and destination. The resulting challenge is to determine what search behavior is useful next based on the current state and which configuration can realize it under the current observations.

\begin{figure*}[t]
    \centering
    \begin{minipage}{0.49\textwidth}
        \centering
        \includegraphics[width=\linewidth]{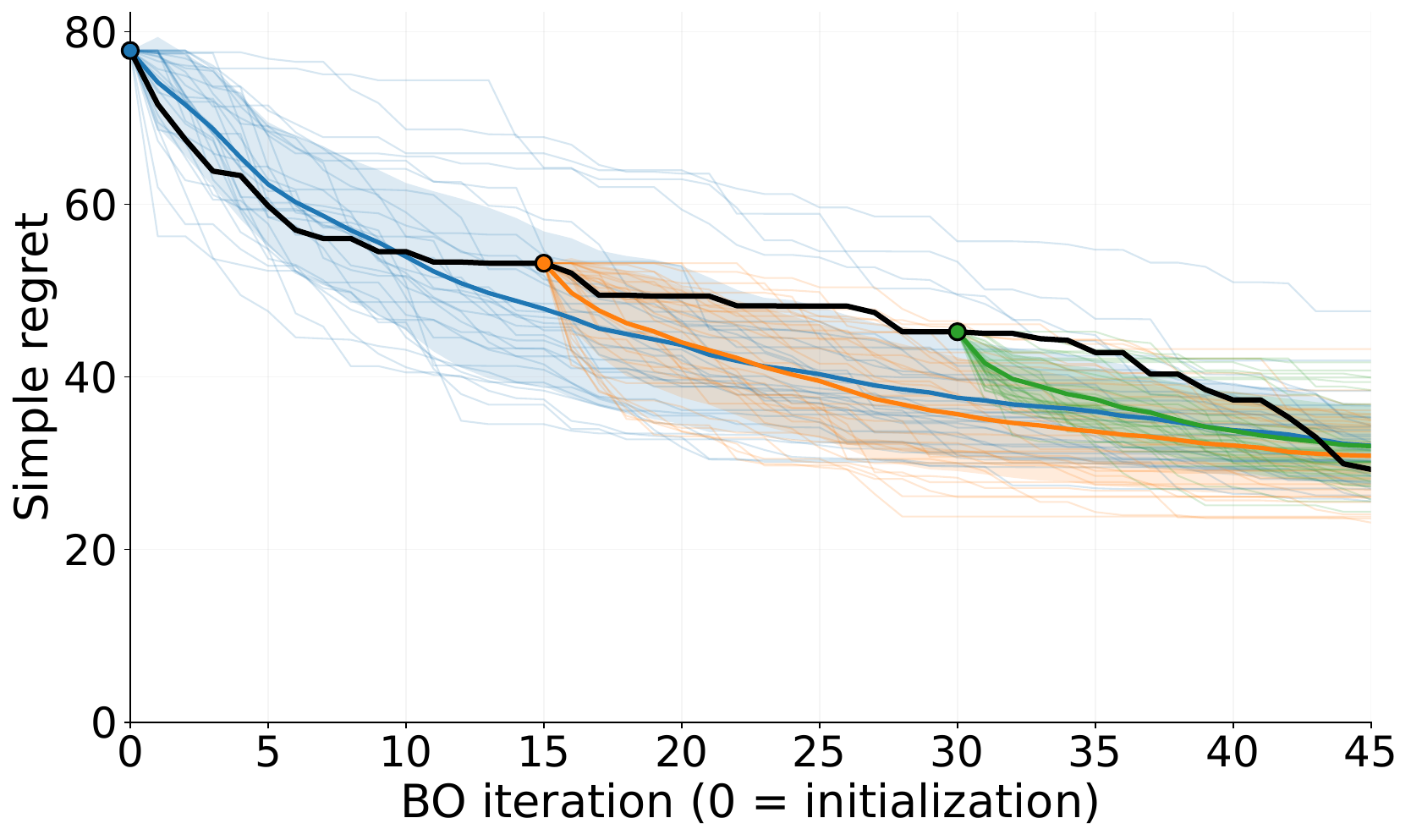}
        \par\small (a) Rastrigin, 5D.
    \end{minipage}
    \hfill
    \begin{minipage}{0.49\textwidth}
        \centering
        \includegraphics[width=\linewidth]{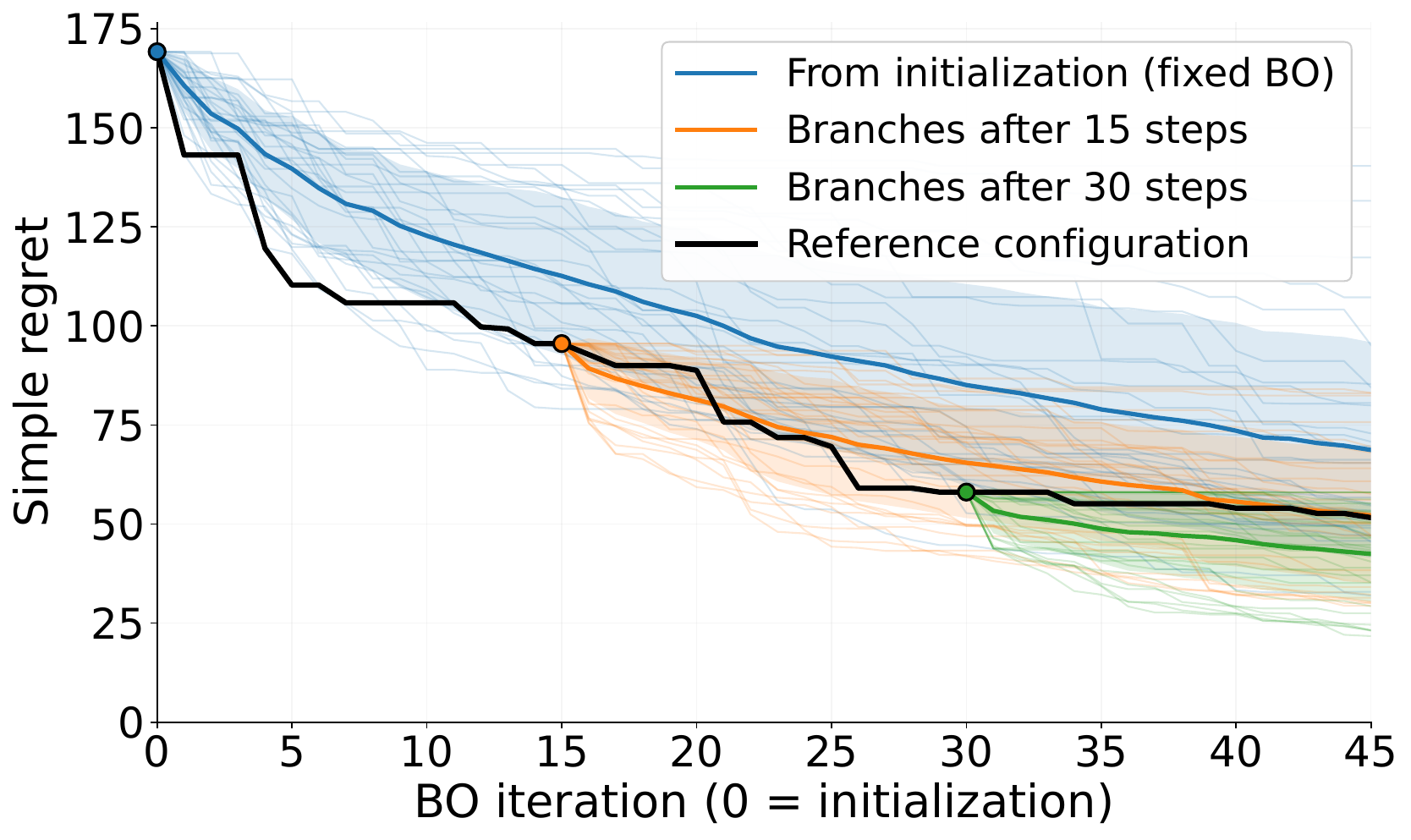}
        \par\small (b) Levy, 15D.
    \end{minipage}
    \caption{Sensitivity to online BO configuration adaptation. \textbf{The benefit of switching BO configurations depends on the problem, switch time, and destination configuration.} Black curves show the unchanged reference configuration. Blue curves use the 29 alternative configurations from initialization; orange and green curves switch to these alternatives after 15 and 30 BO iterations, respectively, inheriting the reference observations collected up to that point. Each alternative configuration is then retained for the remainder of the run. Thin colored curves show individual configurations averaged over ten seeds; thick colored curves and shaded bands show the mean and $\pm$ one standard deviation across these configuration curves. Iteration~0 denotes the state after initialization.
    }
    \label{fig:configuration_sensitivity}
    \vspace{0pt}
\end{figure*}

\section{Method}
\label{section:method}

\subsection{Sequential Meta-Control of Bayesian Optimization}

We formulate BO configuration selection as sequential meta-control over the joint space $\mathcal C = \mathcal G \times \mathcal A$, where $\mathcal G$ and $\mathcal A$ are libraries of surrogate and acquisition profiles, respectively. At iteration $t$, a controller selects the configuration $c_t = (g_t, a_t)$ using memory $\mathcal M_t$ and optional problem context $\mathcal P$. The numerical BO routine then generates the candidate associated with $c_t$:
\begin{equation}
\begin{aligned}
    c_t &= (g_t, a_t) = \Pi(\mathcal M_t; \mathcal P), \\
    \mathbf x_{t+1}^{(c_t)} = \mathcal{BO}_{c_t}(\mathcal D_t) &= \arg\max_{\mathbf x \in \mathcal X}\alpha_{a_t}\! \left(\mathbf x; p_{g_t}(f \mid \mathcal D_t) \right),
\label{eq:bo_candidate}
\end{aligned}
\end{equation}
where $p_g(f \mid \mathcal D_t)$ is the fitted GP posterior for profile $g$ and $\alpha_a$ is the acquisition function specified by $a$. The controller is thus to manage \emph{how} BO searches, not to propose solution points directly.

\subsection{STaSBO: Strategy-guided Hierarchical Control}

STaSBO (Strategy-guided Task-agnostic adaptive Selection for Bayesian Optimization) realizes this meta-control through two coupled stages (Figure~\ref{fig:llm_bo_framework}): \emph{strategy planning} determines the desired search behavior from prior decisions and their outcomes; \emph{strategy-guided configuration selection} identifies a surrogate--acquisition configuration whose prospective candidate supports that behavior. This separation makes search intent explicit before configuration choice and allows it to be retained or revised as new outcomes are observed, while numerical BO remains responsible for candidate generation.

\paragraph{Decision-aware memory.}
STaSBO maintains decision-aware memory $\mathcal M_t$ comprising history memory $\mathcal H_t$ and strategy memory $\mathcal S_t$. $\mathcal H_t = (\mathcal D_t, \mathcal R_t)$ combines accumulated observations $\mathcal D_t$ with completed decision records $\mathcal R_t$, while $\mathcal S_t = (u_t, p_t)$ maintains the trajectory summary $u_t$ and behavioral plans $p_t$. At the start of iteration $t$, the controller receives $\mathcal H_t$ and $\mathcal S_{t-1}$, with $\mathcal S_t$ subsequently produced by \emph{strategy planning}. At the end of iteration $t$, the observed outcome $o_{t+1} = (\mathbf x_{t+1}, y_{t+1})$ and a new decision record $r_{t+1}$ are added to $\mathcal D_t$ and $\mathcal R_t$, respectively, forming $\mathcal H_{t+1} = (\mathcal D_{t+1}, \mathcal R_{t+1})$. At the next iteration, \emph{strategy planning} uses $\mathcal H_{t+1}$ and $\mathcal S_t$ to produce $\mathcal S_{t+1}$, closing the loop by linking prior plans, selected configurations, proposed candidates, and observed outcomes. This preserves the feedback chain \emph{intent $\rightarrow$ configuration $\rightarrow$ candidate $\rightarrow$ outcome}, while keeping control assessments distinct from measured objective improvement.

A new record $r_{t+1} = (\hat p_t, c_t, \rho_t, m_{t+1})$ carries the executed plan $\hat p_t$, selected configuration $c_t$, controller-feedback $\rho_t$, and numerical diagnostics $m_{t+1}$. $\hat p_t$ denotes the behavioral plan associated with the executed action. $\rho_t$ contains the selection rationale and assessment generated before evaluation, while $m_{t+1}$ describes candidate geometry computed against $\mathcal D_t$ and numerical summaries of the observed outcome after evaluation. The decision records are updated as $\mathcal R_{t+1} = \mathcal R_t \cup \{r_{t+1}\}$. Record fields and update conventions are detailed in Appendix~\ref{app:method_details:memory_details}.

\begin{algorithm}[t]
    \caption{STaSBO} 
    \label{alg:stasbo_main}
    \begin{algorithmic}[1]
        \Require $\mathcal D_0$, $\mathcal G$, $\mathcal A$, budget $T$, optional $\mathcal P$, initial $\mathcal S_{-1}$
        \State $\mathcal C \gets \mathcal G \times \mathcal A$; 
        \State $\mathcal R_0 \gets \emptyset$
        \For{$t=0, \ldots, T-1$}
            \State $\mathcal H_t \gets(\mathcal D_t, \mathcal R_t)$
            \State $\mathcal S_t \gets \pi_{\mathrm{plan}}(\mathcal H_t, \mathcal S_{t-1}; \mathcal P)$ \Comment{planning}
            \State $\mathcal E_t\gets\{(g, \phi_{\mathrm{fit}}(\mathcal D_t, g)) : g \in \mathcal G\}$
            \State $\widetilde{\mathcal C}_t \gets \pi_{\mathrm{shortlist}}(\mathcal S_t, \mathcal E_t; \mathcal P, \mathcal C)$ \Comment{shortlisting}
            \ForAll{$c \in \widetilde{\mathcal C}_t$}
                \State $\mathbf x_{t+1}^{(c)} \gets \mathcal{BO}_c(\mathcal D_t)$
                \State $z_t(c) \gets \phi_{\mathrm{diag}}(\mathcal D_t, c, \mathbf x_{t+1}^{(c)})$
            \EndFor
            \State $\mathcal Z_t \gets \{(c, z_t(c)) : c \in \widetilde{\mathcal C}_t\}$
            \State $(c_t, \hat p_t, \rho_t) \gets \pi_{\mathrm{select}}(\mathcal S_t,\widetilde{\mathcal C}_t, \mathcal Z_t;\mathcal P)$ \Comment{selection}
            
            \State $o_{t+1} \gets (\mathbf x_{t+1}^{(c_t)}, f(\mathbf x_{t+1}^{(c_t)}) + \epsilon_{t+1})$
            \State Compute $m_{t+1}$ from $\mathcal D_t$, $z_t(c_t)$, and $o_{t+1}$
            \State $r_{t+1} \gets (\hat p_t,c_t,\rho_t,m_{t+1})$
            \State $\mathcal D_{t+1} \gets \mathcal D_t \cup \{o_{t+1}\}$; $\mathcal R_{t+1} \gets \mathcal R_t \cup \{r_{t+1}\}$
        \EndFor
        \State \Return $\max_{(\mathbf x,y)\in\mathcal D_T} y$ 
    \end{algorithmic}
    \vspace{0pt}
\end{algorithm}

\paragraph{Hierarchical decision rule.}
At each iteration, the controller first updates strategy memory $\mathcal S_t$ using history memory $\mathcal H_t$, then shortlists configurations using fitted-model evidence $\mathcal E_t$, and finally selects among these configurations using prospective candidate diagnostics $\mathcal Z_t$:
\begin{equation}
\begin{aligned}
    \mathcal S_t &= \pi_{\mathrm{plan}}(\mathcal H_t, \mathcal S_{t-1}; \mathcal P), \\
    \widetilde{\mathcal C}_t &= \pi_{\mathrm{shortlist}}(\mathcal S_t, \mathcal E_t; \mathcal P,\mathcal C),\\
    (c_t, \hat p_t, \rho_t)&=\pi_{\mathrm{select}}(\mathcal S_t, \widetilde{\mathcal C}_t, \mathcal Z_t; \mathcal P),
    \quad c_t \in \widetilde{\mathcal C}_t.
\end{aligned}
\label{eq:stasbo_hierarchy}
\end{equation}
Here $\pi_{\mathrm{plan}}$, $\pi_{\mathrm{shortlist}}$, and $\pi_{\mathrm{select}}$ are three LLM decisions; surrogate fitting, candidate generation, and diagnostics are numerical operations. Algorithm~\ref{alg:stasbo_main} summarizes the loop; the expanded implementation appears in Appendix~\ref{app:method_details:overall_algorithm}.

\subsection{Stage 1: Decision-Informed Strategy Planning}

Strategy planning establishes what the next evaluation should attempt by examining both optimization progress and the consequences of previous controller decisions. The planner uses $\mathcal H_t$ to revise $\mathcal S_{t-1}$ to $\mathcal S_t = (u_t, p_t)$ through the first decision in Equation~\eqref{eq:stasbo_hierarchy}. $p_t$ is a set of behavioral plans generated for the current trajectory summary $u_t$, each specifying a target search action, its purpose, preferred candidate properties, and a revision condition, without prescribing a surrogate or acquisition profile. Planning draws on the prior selection rationale $\rho_{t-1}$, recorded candidate geometry from $m_t$, and observed outcome $o_t = (\mathbf x_t, y_t)$. In particular, it distinguishes whether an earlier configuration produced candidate behavior consistent with its associated plan from whether the resulting evaluation improved the objective. These are different signals: an exploratory candidate may satisfy its plan without immediately improving the incumbent. The planner uses both signals to retain or revise its trajectory summary and behavioral plans, maintaining continuity across iterations while adapting search intent to new outcomes. Planning inputs and strategy memory contents are detailed in Appendix~\ref{app:method_details:memory_details}.

\subsection{Stage 2: Strategy-guided Configuration Selection}

While $p_t$ specifies desired search behavior, the candidate proposed by a configuration depends on its surrogate--acquisition pairing and the current observations. STaSBO uses a coarse-to-fine procedure comprising configuration shortlisting and joint selection to resolve this distinction. Shortlisting uses surrogate fit evidence $\mathcal E_t$ to propose plausible configurations for each behavioral plan. Numerical BO then generates a prospective candidate $\mathbf x_{t+1}^{(c)}$ for each distinct shortlisted configuration $c$. Joint selection uses prospective BO-action diagnostics $\mathcal Z_t$ to compare the resulting candidate behaviors with the behavioral plans and select the configuration whose candidate is evaluated next.

\subsubsection{Configuration Shortlisting}
For every $g \in \mathcal G$, $\phi_{\mathrm{fit}}$ fits the surrogate specified by $g$ to $\mathcal D_t$ and extracts learned length scales, observation-noise variance, and kernel output scale. These fitted descriptors are collected as
\begin{equation}
    \mathcal E_t = \{(g, e_t(g)):g\in\mathcal G\}, \quad e_t(g) = \phi_{\mathrm{fit}}(\mathcal D_t, g).
\label{eq:surrogate_fit_evidence}
\end{equation}
The shortlisting decision in Equation~\eqref{eq:stasbo_hierarchy} combines the updated strategy $\mathcal S_t$, fit evidence $\mathcal E_t$, and static profile specifications in a single LLM call. For each behavioral plan in $p_t$, the controller proposes one surrogate profile and a set of associated acquisition profiles. Each acquisition profile specifies a function family and an LLM-proposed parameter value, such as the EI improvement-margin fraction $\xi_{\mathrm{frac}}$ or the UCB coefficient $\kappa$. These values are proposed within predefined continuous ranges. Pairing each plan's surrogate and associated acquisition profiles yields a plan-specific configuration set; their union defines $\widetilde{\mathcal C}_t$, with duplicate configurations merged across plans. Shortlisting and fitted-descriptor definitions are detailed in Appendix~\ref{app:method_details:fit_evidence}, while numerical GP-fitting procedures are described in Appendix~\ref{app:implementation_details:numerical_bo_implementation}.

\begin{table*}[h]
    \centering
    \caption{Results on synthetic mathematical benchmarks. Entries report task-level ranks computed from final simple regret averaged over ten seeds (lower is better). Avg.\ rank is the unweighted mean of the ranks across the twelve function--dimension tasks.}
    \label{tab:main_results_synthetic_rank}
    
    \small
    \setlength{\tabcolsep}{4pt}
    \renewcommand{\arraystretch}{1.12}
    \begin{tabular*}{\linewidth}{@{\extracolsep{\fill}}l*{13}{c}@{}}
        \toprule
        \multirow{2}{*}{Method} &
        \multicolumn{3}{c}{Ackley} & \multicolumn{3}{c}{Levy} & \multicolumn{3}{c}{Rastrigin} & \multicolumn{3}{c}{Rosenbrock} &
        \multirow{2}{*}{\shortstack{Avg.\\rank}} \\
        \cmidrule(lr){2-4}\cmidrule(lr){5-7}\cmidrule(lr){8-10}\cmidrule(lr){11-13}
        & 5D & 8D & 15D & 5D & 8D & 15D & 5D & 8D & 15D & 5D & 8D & 15D & \\
        \midrule
        Random                                 & 12 & 12 & 12 & 12 & 12 & 12 & 12 & 12 & 11 & 12 & 11 & 11 & 11.75 \\
        BO-EI                                  &  4 &  4 &  3 &  8 &  4 &  7 &  9 &  4 &  9 &  8 &  4 &  7 &  5.92 \\
        BO-UCB                                 &  1 &  3 & 11 & 10 & 11 & 11 & 11 & 11 & 12 & 10 & 10 & 10 &  9.25 \\
        \midrule
        GP-Hedge                               & 11 & 10 &  7 &  6 &  3 & 10 &  6 &  1 &  5 &  7 &  6 &  9 &  6.75 \\
        No-PASt-BO                             &  9 &  9 &  9 &  5 &  5 &  9 &  1 & 10 &  8 &  4 &  7 &  5 &  6.75 \\
        TuRBO                                  &  7 &  6 &  2 &  7 &  6 &  1 &  2 &  6 &  6 &  2 &  1 &  3 &  4.08 \\
        SETUP-BO                               & 10 &  8 &  6 &  2 &  9 &  8 &  4 &  9 &  4 &  3 &  3 &  8 &  6.17 \\
        BOOST                                  &  6 &  2 &  4 &  9 &  8 &  4 &  8 &  5 &  7 &  6 &  5 &  2 &  5.50 \\
        \midrule
        LLM                                    &  8 & 11 &  5 & 11 &  7 &  5 &  3 &  3 &  2 & 11 & 12 & 12 &  7.50 \\
        LMABO                                  &  5 &  7 & 10 &  4 &  1 &  6 & 10 &  8 & 10 &  9 &  9 &  4 &  6.92 \\
        CAKE                                   &  3 &  5 &  8 &  1 & 10 &  2 &  5 &  7 &  3 &  5 &  8 &  6 &  5.25 \\
        \midrule
        STaSBO                                 &  2 &  1 &  1 &  3 &  2 &  3 &  7 &  2 &  1 &  1 &  2 &  1 &  2.17 \\
        \bottomrule
    \end{tabular*}
    
    \vspace{0pt}
\end{table*}

\subsubsection{Prospective BO-Action Diagnostics}
For each distinct shortlisted configuration $c \in \widetilde{\mathcal C}_t$, numerical BO generates $\mathbf x_{t+1}^{(c)}$ using Equation~\eqref{eq:bo_candidate}. Without evaluating the objective, $\phi_{\mathrm{diag}}$ records the candidate's coordinates, normalized distances to the incumbent and nearest previously evaluated point, and a distance-based behavior label:
\begin{equation}
\mathcal Z_t = \{(c,z_t(c)):c\in\widetilde{\mathcal C}_t\},
\quad
z_t(c) = \phi_{\mathrm{diag}} \left(\mathcal D_t, c, \mathbf x_{t+1}^{(c)}\right).
\label{eq:prospective_diagnostics}
\end{equation} 
These diagnostics characterize the \emph{actions actually available} under the current observations and provide a common basis for comparing candidate behavior with the behavioral plans. Definitions and normalization details appear in Appendix~\ref{app:method_details:prospective_diagnostics}.

\subsubsection{Joint Selection and Feedback}
The final LLM decision in Equation~\eqref{eq:stasbo_hierarchy} selects one final surrogate-acquisition combination $c_t \in \widetilde{\mathcal C}_t$ by comparing candidate behaviors in $\mathcal Z_t$ with the behavioral plans $p_t$. The accompanying $\rho_t$ records the pre-evaluation selection rationale and assessment, and $\hat{p}_t \in p_t$ records the corresponding plan of the chosen $c_t$. Only the candidate $\mathbf x_{t+1}^{(c_t)}$ is evaluated by the objective. Its observation and the decision record update $\mathcal H_{t+1}$, allowing the next planning step to relate the selected action's intended behavior to its realized geometry and observed search progress. Selection constraints and feedback-record updates are detailed in Appendix~\ref{app:method_details:joint_selection}.

\section{Experiments}
\label{section:experiments}

\subsection{Experimental setup}

\begin{table*}[h]
    \centering
    \caption{Results on scientific and engineering application benchmarks. Entries report task-level ranks computed from final simple regret averaged over ten seeds (lower is better). Avg.\ rank is the unweighted mean of the ranks across the listed tasks.}
    \label{tab:main_results_simulation_rank}
    
    \small
    \setlength{\tabcolsep}{4pt}
    \renewcommand{\arraystretch}{1.12}
    \begin{tabular*}{\linewidth}{@{\extracolsep{\fill}}l*{11}{c}@{}}
        \toprule
        \multirow{2}{*}{Method} & \multirow{2}{*}{AgNP} & \multirow{2}{*}{Concrete} & \multirow{2}{*}{\shortstack{Crossed\\Barrel}} & \multirow{2}{*}{HPLC} & \multirow{2}{*}{P3HT} & \multicolumn{4}{c}{OER Plate} & \multirow{2}{*}{\shortstack{Lunar\\Lander}} & \multirow{2}{*}{\shortstack{Avg.\\rank}} \\ 
        \cmidrule(lr){7-10}
        & & & & & & 3496 & 3851 & 3860 & 4098 & & \\
        \midrule
        Random                                & 11 & 12 & 12 & 11 & 11 & 11 & 11 & 11 & 11 & 11 & 11.2 \\
        BO-EI                                 &  8 &  1 &  4 &  5 &  4 &  8 &  6 &  2 &  7 &  2 &  4.7 \\
        BO-UCB                                &  9 &  3 &  5 &  4 &  1 &  6 &  1 &  1 &  1 & 10 &  4.1 \\
        \midrule
        GP-Hedge                              &  7 &  7 &  7 & 10 &  9 &  9 &  5 &  4 &  4 &  5 &  6.7 \\
        No-PASt-BO                            &  6 &  4 &  6 &  7 &  5 &  1 & 10 &  5 & 10 &  7 &  6.1 \\
        TuRBO                                 &  1 &  6 &  9 &  1 & 10 &  5 &  8 &  7 &  3 &  6 &  5.6 \\
        SETUP-BO                              &  3 &  5 &  8 &  9 &  8 &  7 &  9 &  8 &  9 &  8 &  7.4 \\
        BOOST                                 & 10 & 10 & 10 &  8 &  3 & 10 &  2 & 10 &  8 &  9 &  8.0 \\
        \midrule
        LLM                                   & 12 & 11 & 11 & 12 & 12 & 12 & 12 & 12 & 12 & 12 & 11.8 \\
        LMABO                                 &  5 &  8 &  3 &  2 &  7 &  4 &  3 &  6 &  6 &  1 &  4.5 \\
        CAKE                                  &  4 &  9 &  2 &  6 &  6 &  2 &  7 &  9 &  5 &  3 &  5.3 \\
        \midrule
        STaSBO                                &  2 &  2 &  1 &  3 &  2 &  3 &  4 &  3 &  2 &  4 &  2.6 \\
        \bottomrule
    \end{tabular*}
    
    \vspace{0pt}
\end{table*}

\subsubsection{Benchmarks}
We evaluate STaSBO on 12 synthetic mathematical benchmarks and 10 scientific and engineering application benchmarks spanning different landscapes, domains, and dimensionalities. The mathematical benchmarks comprise Ackley, Levy, Rastrigin, and Rosenbrock~\cite{simulationlib} in 5D, 8D, and 15D, with random shifts that relocate their optima within the search domains. Nine application objectives are deterministic functions constructed by interpolating archived experimental measurements in materials, chemistry, and engineering~\cite{concrete_compressive_strength_165, hase2021olympus, hickman2023olympus}. The Lunar Lander optimizes a controller using a simulator~\cite{eriksson2019scalable}. Benchmark definitions, dataset sources, design variables, constraints, objective construction, and domain context are provided in Appendix~\ref{app:implementation_details:benchmark_definitions}.

\subsubsection{Baselines}
We compare STaSBO with random search, fixed BO configurations, adaptive BO methods, and LLM-guided baselines. Baseline implementations and method-specific settings are provided in Appendix~\ref{app:implementation_details:baselines}.

\paragraph{Fixed BO.} BO-EI and BO-UCB share a Mat\'ern-$5/2$ ARD surrogate profile and use EI with $\xi_{\mathrm{frac}}=0$ and UCB with $\kappa=3$, respectively. These configuration settings remain fixed while GP hyperparameters are refitted as observations accumulate; profile definitions are provided in Appendix~\ref{app:method_details:profile_libraries}.

\paragraph{Adaptive BO.} GP-Hedge~\citep{hoffman2011portfolio} and No-PASt-BO~\citep{vasconcelos2019no} adapt acquisition selection through portfolio reward updates, while SETUP-BO adapts portfolio-selection parameters through Thompson sampling~\citep{vasconcelos2022self}. TuRBO selects candidates through Thompson sampling within an adaptive trust region modeled by a local GP~\citep{eriksson2019scalable}. BOOST jointly selects kernel--acquisition pairs through retrospective BO runs over previously evaluated points, favoring configurations that reach a target value in fewer replay steps~\citep{park2025boost}.

\paragraph{LLM BO.} Among LLM-guided baselines, Direct LLM, labeled LLM in the tables, proposes evaluation points directly from the problem description and observation history. LMABO~\citep{ngo2026adaptive} selects acquisition functions using a structured summary of the optimization state while retaining a prescribed surrogate, whereas CAKE~\citep{suwandi2026adaptive} evolves GP kernels under a prescribed acquisition rule.

\subsubsection{Implementation details}
All LLM-based methods use \texttt{gpt-6-luna} with medium reasoning effort. Methods share five initial observations per task and seed, generated by Latin hypercube sampling for box-constrained domains and $\operatorname{Dirichlet}(\mathbf{1})$ sampling for simplex-constrained domains. Each run performs 45 additional evaluations, one candidate per iteration, giving 50 objective evaluations in total. Within each task, methods are ranked by final simple regret, averaged over ten seeds. Average ranks weight tasks equally within each suite. Normalized results divide each method's seed-mean regret by baseline Random's seed-mean final simple regret on the same task. Initialization, evaluation protocols, and metric definitions are detailed in Appendix~\ref{app:implementation_details:evaluation_protocol}. STaSBO's controller settings, prompt interfaces, and numerical implementation are described in Appendices~\ref{app:implementation_details:controller_implementation}, \ref{app:implementation_details:prompt_formats}, and \ref{app:implementation_details:numerical_bo_implementation}, respectively.

\subsection{Main results}

Tables~\ref{tab:main_results_synthetic_rank} and~\ref{tab:main_results_simulation_rank} show that STaSBO achieves the lowest average rank in both benchmark suites: 2.17 on synthetic mathematical benchmarks and 2.6 on scientific and engineering application benchmarks. On the mathematical suite, STaSBO ranks first on five of the twelve tasks and places in the top two on nine, compared with average ranks of 4.08 for TuRBO and 5.25 for CAKE. On the application suite, STaSBO ranks first on Crossed Barrel, places in the top two on five of the ten tasks, and remains within the top four on every task. Its average rank of 2.60 compares with 4.10 for BO-UCB and 4.50 for LMABO. The relative performance of the fixed BO baselines reverses across suites: BO-EI outperforms BO-UCB on mathematical benchmarks (average ranks of 5.92 versus 9.25), whereas BO-UCB performs better on applications (4.10 versus 4.70) and ranks first on four application tasks. This contrast highlights the problem dependence of fixed configurations and motivates adaptive configuration selection, with STaSBO maintaining strong performance across both suites. Task-wise normalized final regrets are reported in Tables~\ref{tab:main_results_synthetic_regret} and~\ref{tab:main_results_simulation_regret} in the Appendix~\ref{app:additional_results}.

\begin{figure*}[h]
    \centering
    \includegraphics[width=\linewidth]{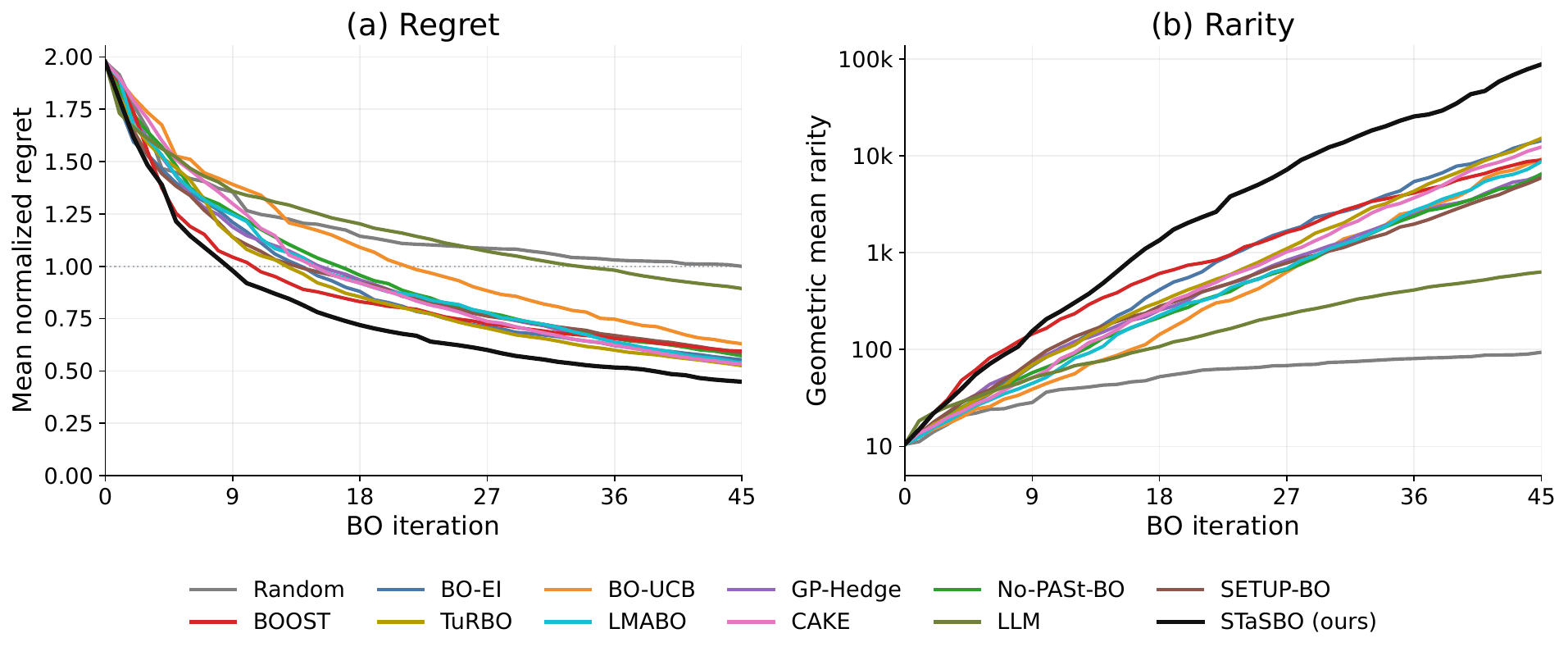}
    \setlength{\abovecaptionskip}{0pt}
    \caption{
    Aggregate optimization performance. \textbf{STaSBO achieves the lowest mean normalized regret and the highest final geometric mean rarity, reaching $5.79 \times$ that of TuRBO, the strongest baseline.} (a) Mean normalized regret across all 22 tasks. Each task's seed-mean regret is divided by Random's mean final regret, then averaged equally across tasks. The dotted line marks Random's reference. (b) Geometric mean incumbent rarity across 21 tasks, excluding Lunar Lander. Rarity is $N = 1/p$, where $p$ is the probability that a uniformly sampled feasible point matches or exceeds the incumbent's objective quality, estimated using the synthetic objectives and application benchmark interpolants. Log rarity is averaged equally over seeds and tasks, then exponentiated; the vertical axis is logarithmic. Lower is better in (a), and higher is better in (b).}
    \label{fig:aggregated_results}
\end{figure*}

Figure~\ref{fig:aggregated_results} complements the rank comparisons with aggregate optimization trajectories and geometric mean rarity. Across all 22 tasks, STaSBO maintains the lowest mean normalized regret from iteration~5 onward, finishing at 0.449 compared with 0.525 for the strongest baseline, TuRBO. Moreover, STaSBO's mean normalized regret falls below 0.75 after 17 BO iterations, compared with 25 for the earliest-reaching baselines, requiring 32\% fewer BO iterations. For the stricter threshold of 0.6, STaSBO requires 28 iterations, compared with 36 for TuRBO, the earliest-reaching baseline, corresponding to a 22.2\% reduction in BO iterations. The rarity analysis provides a complementary perspective by expressing incumbent quality relative to each benchmark’s objective distribution under uniform feasible-domain sampling. For example, rarity $N=10^4$ corresponds to reaching the best 0.01\% of this objective distribution; larger $N$ imposes a stricter quality requirement. Across the 21 benchmarks with rarity estimates, STaSBO reaches a final geometric mean rarity of $87,667$, compared with $15,148$ for TuRBO, a 5.79-fold increase. Thus, in addition to reducing normalized regret, STaSBO finds objective values that are substantially rarer under these reference distributions, as measured by the aggregate geometric mean. Lunar Lander is excluded from the rarity analysis because a reference score distribution is unavailable for its simulator-based objective. Together, these results show lower aggregate regret and attainments of rare, high-quality solutions under the shared evaluation budget. Aggregation, threshold-attainment calculations, and rarity estimation are detailed in Appendix~\ref{app:implementation_details:evaluation_protocol}.

\subsection{Ablation Study}

We examine eight ablations of joint configuration control, planning and memory, and numerical evidence on Ackley-8D. The joint-control variants fix either the surrogate profile (acquisition-only control) or the acquisition (surrogate-only control). The planning and memory variants reconstruct the strategy from observations without prior strategy memory, remove the trajectory summary while retaining behavioral plans and decision feedback, or remove both the planning stage and strategy memory. The evidence variants remove surrogate fit evidence, candidate diagnostics while retaining coordinates, or both diagnostics and coordinates. All conditions use matched initialization and objective-evaluation budgets, with modifications detailed in Appendix~\ref{app:implementation_details:ablations}. Table~\ref{tab:ablation_results} reports results over ten seeds. Full STaSBO achieves a mean final simple regret of 8.25, compared with 9.47--13.89 among the reported variants. Acquisition-only control produces the largest increase in mean regret, and removing candidate diagnostics also increases mean regret even when coordinates remain available. These comparisons support joint configuration control and explicit numerical evidence on this benchmark, while the planning variants compare different combinations of strategy generation and memory.

\begin{table}[h]
    \centering
    \caption{
    Ablations of joint configuration control, planning and memory, and numerical evidence on Ackley-8D. Values report final simple regret as mean $\pm$ sample standard deviation over ten seeds (lower is better).}
    \label{tab:ablation_results}
    \small
    \setlength{\tabcolsep}{3pt}
    \renewcommand{\arraystretch}{1.08}
    \begin{tabular*}{\linewidth}{@{\extracolsep{\fill}}lc@{}}
        \toprule
        Ablations & Regret $\downarrow$ \\
        \midrule
        \textbf{STaSBO (full)}
        & $\mathbf{8.25} \pm 5.75$ \\
        \midrule
        \multicolumn{2}{@{}l}{\textit{Joint Control}} \\
        Acquisition-only control
        & $13.89 \pm 4.42$ \\
        Surrogate-only control
        & $10.94 \pm 6.23$ \\
        \midrule
        \multicolumn{2}{@{}l}{\textit{Planning and Memory}} \\
        Planning from observations
        & $9.47 \pm 7.68$ \\
        w/o trajectory summary $u_t$
        & $11.18 \pm 6.52$ \\
        w/o planning
        & $12.27 \pm 7.55$ \\
        \midrule
        \multicolumn{2}{@{}l}{\textit{Numerical Evidence}} \\
        w/o fit evidence $\mathcal E_t$
        & $12.61 \pm 7.04$ \\
        w/o diagnostics $\mathcal Z_t$ (coords. retained)
        & $11.68 \pm 7.79$ \\
        w/o diagnostics $\mathcal Z_t$ and coords $\mathbf x_{t+1}^{(c)}$.
        & $11.84 \pm 7.01$ \\
        \bottomrule
    \end{tabular*}
\end{table}

\subsection{Analysis}

\begin{table*}[h]
    \centering
    \caption{
    Two consecutive STaSBO decisions on Ackley-8D. Following two intermediate moves without improvement, the controller shifts its focus toward a distinct near-incumbent opportunity and selects a local candidate that improves the incumbent.}
    \label{tab:stasbo_decision_trace}
    
    \small
    \setlength{\tabcolsep}{6pt}
    \renewcommand{\arraystretch}{1.10}
    \begin{tabularx}{\textwidth}{@{}>{\raggedright\arraybackslash}p{0.17\textwidth}>{\raggedright\arraybackslash}X>{\raggedright\arraybackslash}X@{}}
        \toprule
        \textbf{Decision element} & \textbf{Iteration 27} & \textbf{Iteration 28} \\
        \midrule
        
        Trajectory summary $u_t$
        & 31 observations. Incumbent regret is 12.35. A recent intermediate improvement was followed by a miss, leaving the relative value of intermediate and near-incumbent opportunities unresolved.
        & 32 observations. Incumbent regret remains 12.35 after two consecutive intermediate misses. These outcomes do not establish intermediate moves as generally inferior, and refinement evidence remains limited. \\

        \addlinespace[5pt]
        Current focus
        & Compare intermediate and near-incumbent opportunities after a recent intermediate improvement followed by a miss.
        & Assess a distinct near-incumbent opportunity against intermediate alternatives after two consecutive intermediate misses. \\
        
        \addlinespace[5pt]
        Behavioral plans $p_t$
        &
        A: Test whether intermediate movement remains useful after a recent improvement followed by a miss. \textbf{Intermediate opportunity.} \newline
        B: Preserve a distinct local option despite sparse and mixed local evidence. \textbf{Near-incumbent refinement.} \newline
        C: Retain a distant option given limited broad-move evidence. \textbf{Substantially displaced opportunity.}
        &
        A: Retain intermediate opportunities because earlier moves improved. \textbf{Intermediate opportunity.} \newline
        B: Test a distinct near-incumbent point after two consecutive intermediate misses. \textbf{Near-incumbent refinement.} \newline
        C: Retain a distant option given limited broad-move evidence. \textbf{Substantially displaced opportunity.} \\
        
        \addlinespace[5pt]
        Selected action $c_t$
        & Mat\'ern-$1/2$ isotropic + UCB, $\kappa=6.0$.
        & Mat\'ern-$3/2$ isotropic + EI, $\xi_{\mathrm{frac}}=0$. \\

        \addlinespace[5pt]
        Associated plan $\hat p_t$ & Plan A. & Plan B. \\
        
        \addlinespace[5pt]
        Candidate Evidence
        & \textbf{Intermediate move.}\newline $d_{\mathrm{inc}}=d_{\mathrm{near}}=0.0967$.
        & \textbf{Local refinement.}\newline $d_{\mathrm{inc}}=d_{\mathrm{near}}=0.0254$. \\
        
        \addlinespace[5pt]
        Observed outcome
        &
        \textbf{No incumbent improvement.}\newline
        Evaluated Ackley loss: 17.32.\newline
        Incumbent regret: $12.35 \rightarrow 12.35$.
        &
        \textbf{8.96\% reduction in incumbent regret.}\newline
        Evaluated Ackley loss: 11.24.\newline
        Incumbent regret: $12.35 \rightarrow 11.24$.\\
        
        \addlinespace[5pt]
        Subsequent planning
        & Records a second consecutive intermediate miss and shifts the focus toward a distinct near-incumbent opportunity, while retaining other alternatives.
        & Records the local-refinement improvement and retains the near-incumbent focus. Plan B is revised to build on this result. \\
        
        \bottomrule
    \end{tabularx}
    
    \par\vspace{4pt}\noindent
    \begin{minipage}{\textwidth}
        \footnotesize\raggedright
        \textit{Note.} Textual entries are condensed summaries of the LLM generations. Plans are unordered alternatives, and their identifiers do not indicate priority. Numerical values are rounded. $d_{\mathrm{inc}}$ and $d_{\mathrm{near}}$ denote normalized distances to the incumbent and nearest previous observation, respectively. The final row reports planning updates generated at iterations 28 and 29 for the two columns, respectively.
    \end{minipage}
\end{table*}

\paragraph{STaSBO decision. }Table~\ref{tab:stasbo_decision_trace} illustrates how STaSBO connects decision feedback to subsequent configuration choices on Ackley-8D. At iteration 27, the selected configuration produces an intermediate move consistent with its associated plan but does not improve the incumbent. After this second consecutive intermediate miss, planning shifts the current focus toward a distinct near-incumbent opportunity while retaining intermediate and broad alternatives. At iteration 28, the controller changes both the surrogate profile and acquisition setting, selecting a local candidate from a Mat\'ern-$3/2$ isotropic surrogate with EI. Its evaluation reduces incumbent regret from 12.35 to 11.24. This trace illustrates how the controller distinguishes candidate behavior from objective improvement, revises its search focus using recorded outcomes, and connects the revised intent to an available BO action through joint configuration selection.

\begin{figure*}[h]
    \centering
    \includegraphics[width=\linewidth]{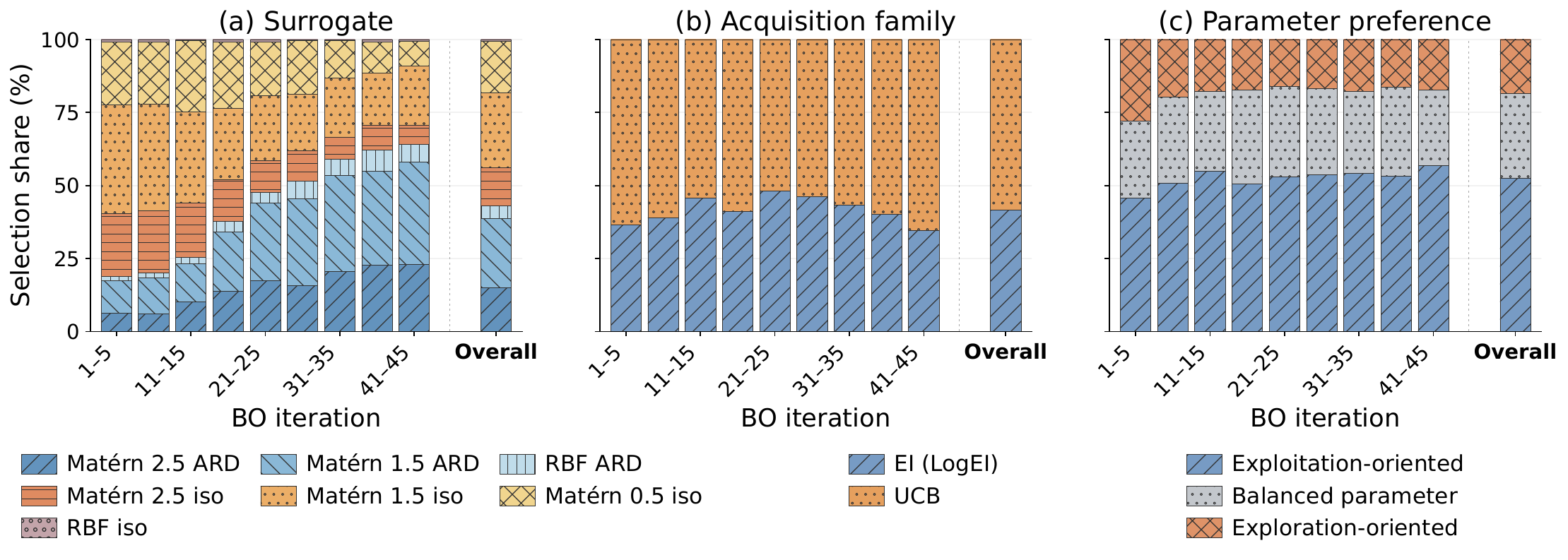}
    \caption{
    Configuration selection patterns across all 22 benchmark tasks. \textbf{Aggregate surrogate preferences shift from isotropic toward ARD profiles over the evaluation budget.} Panels show the selected (a) surrogate profiles, (b) acquisition families, and (c) acquisition-parameter categories. Each bar summarizes final configuration choices within a five-iteration window, averaging over 10 seeds and 22 tasks. ``Overall'' summarizes all 45 BO iterations. Exploitation-oriented: $\xi_{\mathrm{frac}}\leq 0.05$ for EI or $\kappa\leq 1$ for UCB; exploration-oriented: $\xi_{\mathrm{frac}}>0.20$ or $\kappa>3$, respectively. Intermediate values are categorized as balanced.}
    \label{fig:configuration_selection_patterns}
\end{figure*}

\paragraph{Configuration selection patterns.}
Figure~\ref{fig:configuration_selection_patterns} summarizes STaSBO's configuration choices across all 22 benchmark tasks. Surrogate preferences vary substantially over the evaluation budget: isotropic profiles dominate the initial iterations, whereas ARD profiles increase from 18.8\% in iterations 1--5 to 64.1\% in iterations 41--45. Both acquisition families remain active throughout optimization, with UCB accounting for 58.4\% of selections overall. The acquisition-parameter distribution also changes: the share of exploitation-oriented settings rises from 45.8\% in the initial window to 56.8\% in the final window, while exploration-oriented settings decline from 28.0\% to 17.2\%. These patterns show that configuration preferences evolve across optimization stages, involving changes in both surrogate profiles and acquisition parameters. They characterize the controller's choices across the benchmark collection rather than prescribing a common schedule for individual tasks. Selection-record aggregation and parameter-category definitions are detailed in Appendix~\ref{app:implementation_details:selection_patterns}.

\section{Related Work}
\label{section:related_work}

\subsection{Adaptive Bayesian Optimization}
\label{section:related_work:subsection:adaptive_bayesian_optimization}

Meta-learning and prior-based methods for task-adaptive BO learn to decide surrogates or acquisition policies from related tasks or sampled function priors~\citep{volpp2019meta, wistuba2021few, muller2023pfns4bo, wang2024pre}. Their adaptations remain centered on the learned surrogate or policy, making their applicability dependent on source tasks or a training prior. Acquisition portfolios such as GP-Hedge~\citep{hoffman2011portfolio} and No-PASt-BO~\cite{vasconcelos2019no} instead adapt acquisition selection online through numerical reward updates. SETUP-BO further adapts the portfolio's selection parameters~\citep{vasconcelos2022self}. TuRBO adapts the search region using local GP models and trust regions that expand or contract according to observed improvement~\citep{eriksson2019scalable}. BOOST jointly selects kernel--acquisition pairs through retrospective evaluation on accumulated observations~\citep{park2025boost}. STaSBO uses a pretrained LLM to interpret optimization progress, revise search intent, and jointly specify surrogate profiles, acquisition families, and acquisition hyperparameters without BO-specific meta-training. Its configuration decisions are guided by explicit behavioral plans and numerical evidence from the current optimization state.

\subsection{LLM-Based Optimization}
\label{section:related_work:subsection:llm-based_optimization}

OPRO prompts LLMs to propose solution candidates directly~\citep{yang2024large}, while LLAMBO uses LLMs for surrogate modeling and candidate proposal~\citep{liu2024large}. ADO-LLM couples domain-informed LLM proposals with GP-based BO for analog circuit sizing, using high-performing evaluated designs as in-context demonstrations~\citep{yin2024ado}. Hybrid frameworks introduce structured guidance: BORA uses revisable hypotheses to propose solutions~\citep{cisse2025language}, LLINBO integrates LLM proposals with BO-based assessment~\citep{chang2025llinbo}, and Reasoning BO selects BO-generated candidates using accumulated evidence~\citep{yang2025reasoning}. Separating planning from proposal generation has been studied in multi-agent LLM optimization~\citep{carbonati2026multi}. For component adaptation, CAKE evolves kernels under a prescribed acquisition function~\citep{suwandi2026adaptive}, whereas LMABO selects acquisition functions under a prescribed surrogate family~\citep{ngo2026adaptive}. STaSBO couples \emph{strategy planning} with \emph{strategy-guided configuration selection} through a decision-aware \emph{memory system} linking previous plans and configuration decisions to observed outcomes. Within this loop, \emph{surrogate fit evidence} supports plan-conditioned configuration shortlisting, and \emph{prospective BO-action diagnostics} enable comparison of the resulting numerical candidates with those plans. This connects revisable search intentions to the candidate behaviors available under the current observations.

\section{Conclusion and Discussion}
\label{section:conclusion}

STaSBO is a task-agnostic LLM controller for Bayesian optimization that jointly selects surrogate–acquisition configurations through strategy planning and strategy-guided configuration selection. The memory system supports ongoing revision of the search strategy, while surrogate fit evidence and prospective BO-action diagnostics tie that strategy to the candidates BO actually proposes. Across synthetic mathematical and scientific application benchmarks, STaSBO achieves strong aggregate performance without requiring a well-suited configuration to be identified in advance.

More broadly, this framework treats the BO configuration as part of the search itself, with each decision informed by accumulated evidence, interpretations of progress, and the behavior of available configurations. The LLM contributes contextual reasoning, while numerical BO retains modeling and candidate generation. This division of roles provides a framework for adapting BO to expensive scientific and engineering design problems through strategic configuration control grounded in numerical evidence.

\bibliography{llm_bo}
\bibliographystyle{icml2026}

\newpage
\appendix
\onecolumn

\section{Sensitivity Analysis Details}
\label{app:sensitivity_analysis}

\subsection{Benchmarks and Shared Evaluation Protocol}
\label{app:sensitivity_analysis:protocol}

For sensitivity analysis in Section~\ref{section:background:subsection:sensitivity}, we evaluate three synthetic mathematical functions (Ackley, Levy, and Rastrigin) in 5D and 15D, yielding six problem settings. Table~\ref{tab:app_main_synthetic_domains} summarizes their domains and random shifts. For each seed, we evaluate $h(\mathbf{x}-\boldsymbol{\delta})$, sampling the shift independently across coordinates so that the minimizer lies uniformly within the central 80\% of the search box. All objectives are deterministic with minimum value zero; BO internally maximizes the negative objective. Each run starts from five Latin hypercube samples and performs 45 additional evaluations, one per BO iteration. We use ten seeds, numbered 0--9, sharing the objective shift and initial observations across configurations within each problem and seed. Final simple regret is computed from the best objective value observed so far, including initialization. Iteration~0 denotes the incumbent after the five initial evaluations.

\subsection{Configuration Pool and Numerical BO}
\label{app:sensitivity_analysis:configurations}

The sensitivity analysis uses the Cartesian product of the five surrogate and six acquisition profiles in Table~\ref{tab:sensitivity_profiles}, giving 30 configurations. This finite diagnostic pool is defined separately from STaSBO's parameterized configuration space.

\begin{table}[h]
    \centering
    \caption{Surrogate and acquisition profiles used in the configuration sensitivity analysis. Crossing S1--S5 with A1--A6 gives 30 configurations. ARD uses a separate length scale for each input dimension; isotropic kernels use one shared length scale across dimensions.}
    \label{tab:sensitivity_profiles}
    \small
    \setlength{\tabcolsep}{5pt}
    \renewcommand{\arraystretch}{1.12}
    \begin{tabular*}{\linewidth}{@{\extracolsep{\fill}}lll@{}}
        \toprule
        Profile & Kernel / acquisition & Specification \\
        \midrule
        \multicolumn{3}{@{}l}{\textit{Surrogate profiles}} \\
        \addlinespace[2pt]
        S1 & Mat\'ern-$5/2$ & ARD \\
        S2 & Mat\'ern-$5/2$ & Isotropic \\
        S3 & Mat\'ern-$3/2$ & ARD \\
        S4 & RBF            & ARD \\
        S5 & RBF            & Isotropic \\
        \midrule
        \multicolumn{3}{@{}l}{\textit{Acquisition profiles}} \\
        \addlinespace[2pt]
        A1 & Expected improvement & $\xi_{\mathrm{frac}} = 0.00$ \\
        A2 & Expected improvement & $\xi_{\mathrm{frac}} = 0.05$ \\
        A3 & Expected improvement & $\xi_{\mathrm{frac}} = 0.20$ \\
        A4 & Upper confidence bound & $\kappa = 0.5$ \\
        A5 & Upper confidence bound & $\kappa = 3.0$ \\
        A6 & Upper confidence bound & $\kappa = 6.0$ \\
        \bottomrule
    \end{tabular*}
\end{table}

For EI, the coefficient $\xi_{\mathrm{frac}}$ determines the improvement margin relative to the observed score range:
\begin{equation}
\xi_t = \xi_{\mathrm{frac}} \left( \max_{(\mathbf{x}, y) \in \mathcal D_t} y - \min_{(\mathbf{x}, y) \in \mathcal D_t} y \right).
\label{eq:sensitivity_ei_margin}
\end{equation}
UCB is computed as $\mu_t(\mathbf{x}) + \kappa \sigma_t(\mathbf{x})$. A fixed configuration retains its surrogate profile, acquisition family, and acquisition parameter throughout the run. GP parameters are refitted as observations accumulate, and the effective EI margin changes with the observed score range.

GPs are fitted in double precision using an exact marginal-likelihood objective, with learned length scales, kernel output scale, and observation-noise variance. Input normalization uses the coordinate-wise ranges of the current training observations, and output standardization uses the observed scores. Acquisition refinement and duplicate exclusion use unit-box coordinates defined by the search-domain bounds. At each iteration, acquisition optimization uses a uniformly sampled pool of 10,000 points, followed by continuous refinement from the 20 highest-scoring eligible points, with an optimizer iteration limit of 500. EI is optimized in logarithmic form for numerical stability. Points within a maximum-coordinate distance of $10^{-6}$ from an existing observation in unit-box coordinates are excluded.

\subsection{Oracle Selection and Cross-Problem Transfer}
\label{app:sensitivity_analysis:transfer}

For each problem (function $\times$ dimension), we rank the 30 fixed configurations in Table~\ref{tab:sensitivity_profiles} by their mean final simple regret over ten seeds, assigning average ranks to exact ties. The oracle configuration is the pair with the lowest mean final simple regret on its source problem; the six source problems have unique oracle configurations. Table~\ref{tab:configuration_transfer} reports each source oracle's rank on every target problem, using that target's independently executed fixed-configuration runs. Table~\ref{tab:app_fixed_configuration_results} gives the regret values underlying these ranks. Oracle selection is retrospective and uses completed runs; it provides a diagnostic reference for configuration sensitivity.

\begin{table*}[h]
    \centering
    \caption{Final simple regret of the 30 fixed BO configurations, grouped by surrogate. Task entries give the mean over ten seeds, with the within-task rank in parentheses. Bold entries identify each task's oracle. Mean rank averages the six task ranks. Acquisition settings give the EI margin fraction $\xi_{\mathrm{frac}}$ or the UCB coefficient $\kappa$.}
    \label{tab:app_fixed_configuration_results}
    \small
    \setlength{\tabcolsep}{2.5pt}
    \renewcommand{\arraystretch}{1.03}
        \begin{tabular*}{\linewidth}{@{\extracolsep{\fill}}ll*{7}{r}@{}}
        \toprule
         & & \multicolumn{2}{c}{Ackley} & \multicolumn{2}{c}{Levy} & \multicolumn{2}{c}{Rastrigin} & \\
        \cmidrule(lr){3-4}\cmidrule(lr){5-6}\cmidrule(lr){7-8}
        Surrogate & Acquisition & 5D & 15D & 5D & 15D & 5D & 15D & Mean rank \\
        \midrule
        \multirow{6}{*}{\shortstack[l]{\textbf{S1}\\Mat\'ern-$5/2$\\ARD}} & A1\enspace EI (0.00) & 7.312 (17) & 19.299 (13) & 4.852 (14) & 56.759 (16) & 31.130 (15) & 178.093 (18) & 15.50 \\
         & A2\enspace EI (0.05) & 6.231 (12) & 20.163 (22) & 4.968 (16) & 69.683 (19) & 30.215 (12) & 176.210 (16) & 16.17 \\
         & A3\enspace EI (0.20) & 6.770 (13) & 20.565 (28) & 7.675 (25) & 84.520 (24) & 41.888 (29) & 202.348 (22) & 23.50 \\
         & A4\enspace UCB (0.5) & 12.466 (27) & 19.134 (10) & \textbf{2.296 (1)} & 51.313 (9) & 30.209 (11) & 162.898 (6) & 10.67 \\
         & A5\enspace UCB (3.0) & \textbf{3.221 (1)} & 20.421 (26) & 5.881 (22) & 85.357 (25) & 34.348 (24) & 229.786 (27) & 20.83 \\
         & A6\enspace UCB (6.0) & 10.204 (25) & 20.848 (30) & 9.864 (29) & 131.593 (29) & 35.364 (26) & 269.739 (30) & 28.17 \\
        \midrule
        \multirow{6}{*}{\shortstack[l]{\textbf{S2}\\Mat\'ern-$5/2$\\isotropic}} & A1\enspace EI (0.00) & 5.963 (11) & 19.787 (16) & 3.294 (6) & 53.401 (13) & 26.273 (3) & \textbf{143.724 (1)} & 8.33 \\
         & A2\enspace EI (0.05) & 5.435 (6) & 19.295 (12) & 4.828 (13) & 47.184 (6) & 28.544 (9) & 163.626 (8) & 9.00 \\
         & A3\enspace EI (0.20) & 6.902 (14) & 17.423 (6) & 5.486 (19) & 70.794 (20) & 35.222 (25) & 172.337 (12) & 16.00 \\
         & A4\enspace UCB (0.5) & 13.326 (29) & 19.758 (15) & 5.812 (21) & 54.959 (15) & 25.833 (2) & 167.507 (9) & 15.17 \\
         & A5\enspace UCB (3.0) & 4.783 (3) & 16.894 (4) & 5.104 (17) & 32.909 (2) & 27.670 (6) & 159.524 (5) & 6.17 \\
         & A6\enspace UCB (6.0) & 5.297 (5) & 18.928 (9) & 9.734 (28) & 66.240 (18) & 29.287 (10) & 163.029 (7) & 12.83 \\
        \midrule
        \multirow{6}{*}{\shortstack[l]{\textbf{S3}\\Mat\'ern-$3/2$\\ARD}} & A1\enspace EI (0.00) & 5.829 (10) & 19.868 (17) & 2.300 (2) & 45.529 (3) & 33.123 (20) & 180.052 (20) & 12.00 \\
         & A2\enspace EI (0.05) & 8.708 (22) & 19.466 (14) & 4.013 (8) & 72.882 (21) & 34.204 (23) & 178.653 (19) & 17.83 \\
         & A3\enspace EI (0.20) & 5.797 (9) & 20.096 (21) & 7.804 (26) & 80.557 (23) & 38.378 (28) & 214.071 (24) & 21.83 \\
         & A4\enspace UCB (0.5) & 16.103 (30) & 18.614 (8) & 3.089 (4) & 52.550 (11) & 32.422 (18) & 172.843 (13) & 14.00 \\
         & A5\enspace UCB (3.0) & 5.709 (7) & 20.205 (23) & 6.798 (23) & 95.190 (26) & 33.849 (22) & 227.999 (26) & 21.17 \\
         & A6\enspace UCB (6.0) & 7.546 (20) & 20.624 (29) & 13.073 (30) & 117.297 (28) & 47.615 (30) & 266.137 (28) & 27.50 \\
        \midrule
        \multirow{6}{*}{\shortstack[l]{\textbf{S4}\\RBF\\ARD}} & A1\enspace EI (0.00) & 7.058 (16) & 20.039 (19) & 3.070 (3) & 47.176 (5) & 32.806 (19) & 173.958 (14) & 12.67 \\
         & A2\enspace EI (0.05) & 7.337 (18) & 19.887 (18) & 3.768 (7) & 65.431 (17) & 33.335 (21) & 177.087 (17) & 16.33 \\
         & A3\enspace EI (0.20) & 8.755 (23) & 20.274 (24) & 7.353 (24) & 79.943 (22) & 36.666 (27) & 205.132 (23) & 23.83 \\
         & A4\enspace UCB (0.5) & 13.068 (28) & 19.192 (11) & 4.520 (11) & 54.660 (14) & 26.975 (4) & 171.476 (11) & 13.17 \\
         & A5\enspace UCB (3.0) & 7.875 (21) & 20.408 (25) & 5.116 (18) & 107.218 (27) & 31.456 (16) & 220.250 (25) & 22.00 \\
         & A6\enspace UCB (6.0) & 10.301 (26) & 20.477 (27) & 7.886 (27) & 140.389 (30) & 30.242 (13) & 268.499 (29) & 25.33 \\
        \midrule
        \multirow{6}{*}{\shortstack[l]{\textbf{S5}\\RBF\\isotropic}} & A1\enspace EI (0.00) & 3.875 (2) & 17.698 (7) & 4.306 (10) & \textbf{31.949 (1)} & 27.072 (5) & 152.158 (2) & 4.50 \\
         & A2\enspace EI (0.05) & 7.375 (19) & 16.030 (2) & 4.105 (9) & 49.638 (8) & 31.570 (17) & 168.652 (10) & 10.83 \\
         & A3\enspace EI (0.20) & 7.005 (15) & \textbf{15.523 (1)} & 4.719 (12) & 45.751 (4) & 30.430 (14) & 155.929 (3) & 8.17 \\
         & A4\enspace UCB (0.5) & 9.821 (24) & 20.064 (20) & 4.903 (15) & 48.991 (7) & 28.529 (8) & 180.757 (21) & 15.83 \\
         & A5\enspace UCB (3.0) & 4.786 (4) & 16.454 (3) & 3.277 (5) & 52.640 (12) & \textbf{25.582 (1)} & 157.728 (4) & 4.83 \\
         & A6\enspace UCB (6.0) & 5.792 (8) & 17.009 (5) & 5.793 (20) & 51.621 (10) & 27.946 (7) & 175.525 (15) & 10.83 \\
        \bottomrule
    \end{tabular*}
\end{table*}

\subsection{Configuration Switching Protocol}
\label{app:sensitivity_analysis:switching}

We evaluate one-time configuration changes on all six sensitivity tasks. Figure~\ref{fig:configuration_sensitivity} shows the trajectories for Rastrigin-5D and Levy-15D. For each problem, the reference is the configuration ranked tenth by mean final simple regret within the 30-configuration pool: Mat\'ern-$5/2$ isotropic with UCB ($\kappa = 6$) (S2-A6) for Rastrigin-5D, and RBF isotropic with UCB ($\kappa = 6$) (S5-A6) for Levy-15D.

After 0 (initialization), 15 or 30 completed BO iterations, each reference trajectory is branched to the other 29 configurations. Every branch inherits the reference observations available at that checkpoint and retains its assigned configuration for the remaining evaluations. A branch after 15 iterations therefore inherits 20 observations and performs 30 further evaluations; a branch after 30 iterations inherits 35 observations and performs 15 further evaluations. All resulting trajectories contain 50 evaluations, including the shared 5 initialization. For each configuration and switching point, we first average the regret trajectory over the same ten seeds. Thin colored curves show these configuration-specific means. Thick colored curves average the 29 alternative configuration curves with equal weights, and shaded bands show $\pm$ one population standard deviation across those curves. Thus, the bands describe variation across configurations, rather than seed-level uncertainty or confidence intervals. The black curve shows the ten-seed mean of the unchanged reference.

\begin{table}[h]
    \centering
    \caption{Number of alternative configurations, out of 29, that achieve lower mean final simple regret than continuing with the reference configuration. Each comparison uses the final regret at iteration 45, averaged over the same ten seeds.}
    \label{tab:sensitivity_switch_outcomes}
    \small
    \begin{tabular}{@{}lcc@{}}
        \toprule
         & \multicolumn{2}{c}{Switch after} \\
        \cmidrule(lr){2-3}
        Task & 15 iterations & 30 iterations \\
        \midrule
        Ackley-5D     &  3 &  8 \\
        Ackley-15D    & 13 & 12 \\
        Levy-5D       & 11 & 23 \\
        Levy-15D      & 15 & 20 \\
        Rastrigin-5D  & 13 &  8 \\
        Rastrigin-15D &  9 &  3 \\
        \bottomrule
    \end{tabular}
\end{table}

The effect of switch timing varies by task: delaying the switch from iteration 15 to 30 increases the number of alternatives that outperform the reference from 11 to 23 on Levy-5D, but reduces it from 9 to 3 on Rastrigin-15D (Table~\ref{tab:sensitivity_switch_outcomes}).

\section{Method Details}
\label{app:method_details}

\subsection{Detailed Overall Algorithm}
\label{app:method_details:overall_algorithm}

\begin{algorithm}[h]
\caption{STaSBO: detailed two-stage BO configuration control}
\label{alg:stasbo}
\small
\begin{algorithmic}[1]
\Require Initial observations $\mathcal D_0$; surrogate library $\mathcal G$; acquisition space $\mathcal A$; additional evaluation budget $T$
\Require Optional context $\mathcal P$; initial strategy memory $\mathcal S_{-1}$
\State $\mathcal R_0 \gets \emptyset$; $\mathcal C \gets \mathcal G \times \mathcal A$
\State Initialize previous-generation feedback as empty
\For{$t=0, \ldots, T-1$}
    \State $\mathcal H_t \gets (\mathcal D_t, \mathcal R_t)$
    \State Construct the shared decision context and recent-record view
    \State $\mathcal S_t \gets \pi_{\mathrm{plan}}(\mathcal H_t, \mathcal S_{t-1}; \mathcal P)$
    \State $\mathcal E_t \gets \emptyset$
    \ForAll{$g \in \mathcal G$}
        \State $e_t(g) \gets \phi_{\mathrm{fit}}(\mathcal D_t, g)$
        \State Cache the fitted GP if available; otherwise record fit unavailability
        \State $\mathcal E_t \gets \mathcal E_t \cup \{(g, e_t(g))\}$
    \EndFor
    \State Stop without objective evaluation if no GP fit is available
    \State $\widetilde{\mathcal C}_t \gets \pi_{\mathrm{shortlist}}(\mathcal S_t, \mathcal E_t; \mathcal P, \mathcal C)$
    \State Sample a shared pool for continuous acquisition optimization
    \ForAll{$c \in \widetilde{\mathcal C}_t$}
        \State Attempt to generate and cache $\mathbf x_{t+1}^{(c)} \gets \mathcal{BO}_c(\mathcal D_t)$ using the fitted GP and shared pool
        \State Record any candidate-generation failure
    \EndFor
    \State Retain in $\widetilde{\mathcal C}_t$ only configurations with valid cached candidates
    \State Stop without objective evaluation if $\widetilde{\mathcal C}_t = \emptyset$
    \State $\mathcal Z_t \gets \{ (c, \phi_{\mathrm{diag}}(\mathcal D_t, c, \mathbf x_{t+1}^{(c)})): c \in \widetilde{\mathcal C}_t \}$
    \State Group identical candidate coordinates for presentation, retaining all generating configurations
    \State $(c_t, \hat p_t, \rho_t) \gets \pi_{\mathrm{select}}(\mathcal S_t, \widetilde{\mathcal C}_t, \mathcal Z_t; \mathcal P)$
    \State Retain the accompanying plan association $\hat p_t$, which may be absent
    \State $\mathbf x_{t+1} \gets \mathbf x_{t+1}^{(c_t)}$
    \State Evaluate the cached point and observe $y_{t+1}$
    \State $o_{t+1} \gets (\mathbf x_{t+1}, y_{t+1})$
    \State Compute numerical record fields $m_{t+1}$ using $o_{t+1}$ and the pre-evaluation context
    \State $r_{t+1} \gets (\hat p_t, c_t, \rho_t, m_{t+1})$
    \State $\mathcal D_{t+1} \gets \mathcal D_t \cup \{ o_{t+1} \}$
    \State $\mathcal R_{t+1} \gets \mathcal R_t \cup \{ r_{t+1} \}$
    \State Retain generation diagnostics, failures, and the selected configuration for next-iteration feedback
\EndFor
\State \Return $\displaystyle \max_{(\mathbf x, y) \in \mathcal D_T} y$
\end{algorithmic}
\end{algorithm}

Algorithm~\ref{alg:stasbo} expands the two-stage controller into three LLM decisions: strategy planning, configuration shortlisting, and joint configuration selection. The LLM decision functions retain the compact notation of the main text and include prompt construction, response validation, and deterministic postprocessing. Table~\ref{tab:app_llm_interfaces} specifies the information provided to each LLM call. All calls receive the task environment, calculated current context, and a bounded view of recent decision records. The complete observation set $\mathcal D_t$ remains available for numerical BO; it is not supplied as a complete raw observation table to the LLM. An iteration has three LLM decisions, although invalid responses can require additional LLM calls.

\begin{table}[h]
    \centering
    \small
    \setlength{\tabcolsep}{5pt}
    \renewcommand{\arraystretch}{1.12}
    \caption{Inputs and outputs of the three LLM decision functions. All functions additionally receive the task environment, calculated current context, and bounded recent decision records. The environment includes the optional problem description, search-space specification, constraints, and remaining evaluation budget.}
    \label{tab:app_llm_interfaces}
        \begin{tabularx}{\linewidth}{@{}p{0.18\linewidth}XX@{}}
        \toprule
        Decision & Stage-specific inputs & Output \\
        \midrule
        Strategy planning &
        Previous strategy memory $\mathcal S_{t-1}$ and aggregate feedback from the preceding candidate-generation step. &
        Updated strategy memory $\mathcal S_t$, comprising the trajectory summary and alternative behavioral plans. \\
        \addlinespace[4pt]
        Configuration shortlisting & 
        Current strategy $\mathcal S_t$, fixed surrogate specifications, fit evidence $\mathcal E_t$, acquisition families and parameter ranges, and previous-generation feedback. &
        One surrogate and associated parameterized acquisitions per plan. Validation, union, and deduplication yield $\widetilde{\mathcal C}_t$, retaining originating-plan associations. \\
        \addlinespace[4pt]
        Joint configuration selection &
        Current strategy $\mathcal S_t$, static definitions of the retained surrogates, and current candidate diagnostics $\mathcal Z_t$ with their generating configurations. &
        Selected configuration $c_t$, an optional associated plan $\hat p_t$, and pre-evaluation feedback $\rho_t$ containing the rationale and advisory alignment assessment. \\
        \bottomrule
    \end{tabularx}
\end{table}

The shortlisting operation validates the plan-specific configurations and merges configurations with identical surrogate profiles, acquisition families, and parameter values. Each distinct configuration is optimized once. Subsequent grouping of exactly identical candidate coordinates changes their presentation to the selector while preserving every generating configuration. The final selector chooses a valid configuration associated with a cached candidate; the candidate is not regenerated after selection.

The associated plan $\hat p_t$ accompanies $(c_t, \rho_t)$ in the selector's structured output. It may refer to any current behavioral plan or be absent, independently of the configuration's originating plans. The feedback $\rho_t$ records the pre-evaluation rationale and advisory plan-alignment assessment. Candidate geometry is computed against $\mathcal D_t$ before evaluation; the subsequent observation supplies the outcome fields in $m_{t+1}$. The resulting record links the selected plan, configuration, candidate behavior, and observed outcome.

\subsection{Configuration Space and Profile Parameterization}
\label{app:method_details:profile_libraries}

A BO configuration $c = (g, a) \in \mathcal C = \mathcal G \times \mathcal A$ combines a surrogate profile $g$ with an acquisition specification $a$. The surrogate library $\mathcal G$ contains the seven profiles in Table~\ref{tab:app_surrogate_profiles}. Each profile fixes the kernel family, Mat\'ern smoothness parameter where applicable, and length-scale parameterization. Length scales, kernel output scale, and observation-noise variance are fitted to the current observations rather than fixed hyperparameters.

\begin{table}[h]
    \centering
    \small
    \setlength{\tabcolsep}{5pt}
    \renewcommand{\arraystretch}{1.12}
    \caption{Surrogate profiles available to STaSBO. ARD uses one learned length scale per input dimension; isotropic profiles use one shared length scale. All profiles use a learned kernel output scale and learned observation-noise variance.}
    \label{tab:app_surrogate_profiles}
    \begin{tabular*}{\linewidth}{@{\extracolsep{\fill}}lll@{}}
        \toprule
        Profile & Kernel & Length-scale parameterization \\
        \midrule
        $g_1$ & Mat\'ern-$5/2$ & ARD \\
        $g_2$ & Mat\'ern-$5/2$ & Isotropic \\
        $g_3$ & Mat\'ern-$3/2$ & ARD \\
        $g_4$ & RBF            & ARD \\
        $g_5$ & RBF            & Isotropic \\
        $g_6$ & Mat\'ern-$3/2$ & Isotropic \\
        $g_7$ & Mat\'ern-$1/2$ & Isotropic \\
        \bottomrule
    \end{tabular*}
\end{table}

An acquisition specification identifies a family and its search parameter value. STaSBO uses expected improvement (EI) and upper confidence bound (UCB), with the ranges in Table~\ref{tab:app_acquisition_profiles}. The acquisition space is
\begin{equation}
\mathcal A = \bigl( \{\mathrm{EI}\} \times [0,1] \bigr) \cup \bigl( \{\mathrm{UCB}\} \times [0,20] \bigr),
\label{eq:app_acquisition_space}
\end{equation}
where the EI coordinate is $\xi_{\mathrm{frac}}$ and the UCB coordinate is $\kappa$. The LLM proposes these parameter values during configuration shortlisting.

\begin{table}[h]
    \centering
    \small
    \setlength{\tabcolsep}{5pt}
    \renewcommand{\arraystretch}{1.12}
    \caption{Acquisition families and parameter ranges available to STaSBO. Parameters are proposed as numerical values within the stated intervals.}
    \label{tab:app_acquisition_profiles}
    \begin{tabular*}{\linewidth}{@{\extracolsep{\fill}}lll@{}}
        \toprule
        Acquisition family & Parameter & Allowed range \\
        \midrule
        Expected improvement & $\xi_{\mathrm{frac}}$ & $[0, 1]$ \\
        Upper confidence bound & $\kappa$ & $[0, 20]$ \\
        \bottomrule
    \end{tabular*}
\end{table}

Under the internal maximization convention, let
\[
b_t = \max_{(\mathbf x, y) \in \mathcal D_t} y,
\qquad
\xi_t = \xi_{\mathrm{frac}} \left( \max_{(\mathbf x, y) \in \mathcal D_t} y - \min_{(\mathbf x, y) \in \mathcal D_t} y \right).
\]
The corresponding acquisition functions are
\begin{equation}
\begin{aligned}
\alpha_{\mathrm{EI}, \xi_{\mathrm{frac}}} \! \left( \mathbf x; p_g(f \mid \mathcal D_t) \right)
&= \mathbb E_{p_g(f \mid \mathcal D_t)} \left[ (f(\mathbf x) - b_t - \xi_t)_+\right], \\
\alpha_{\mathrm{UCB}, \kappa} \! \left( \mathbf x; p_g(f \mid \mathcal D_t) \right)
&= \mu_{g, t}(\mathbf x) + \kappa \sigma_{g, t}(\mathbf x),
\end{aligned}
\label{eq:app_acquisition_definitions}
\end{equation}
where $(v)_+ = \max \{ v, 0 \}$, and $\mu_{g, t}$ and $\sigma_{g, t}$ are the posterior mean and standard deviation of the fitted surrogate. Thus, $\xi_{\mathrm{frac}}$ is a fraction of the observed score range, rather than an absolute improvement margin. EI is optimized in logarithmic form for numerical stability. UCB uses $\kappa$ as the multiplier of posterior standard deviation; its BoTorch implementation sets $\beta = \kappa^2$. At $\kappa = 0$, UCB reduces to the posterior mean.

Each proposed acquisition parameter is held fixed while its acquisition function is optimized over $\mathbf x\in\mathcal X$. For the main experiments, each of three behavioral plans nominates one surrogate and three acquisition specifications. The plan-specific configuration sets are combined and deduplicated, yielding at most nine distinct configurations per iteration. Plans may share surrogate profiles and acquisition settings; no fixed allocation between EI and UCB is imposed. The finite configuration pool used for the sensitivity analysis is specified separately in Appendix~\ref{app:sensitivity_analysis:configurations}.

\subsection{Memory Contents and Decision Context}
\label{app:method_details:memory_details}

\paragraph{History memory.}
History memory $\mathcal H_t = (\mathcal D_t, \mathcal R_t)$ retains all observations and completed controller decisions. Initialization observations belong to $\mathcal D_0$ and have no preceding controller decision, so $\mathcal R_0 = \emptyset$. At iteration $t$, the selected configuration produces observation $o_{t+1}$ and decision record
\[
r_{t+1} = (\hat p_t,c_t,\rho_t,m_{t+1}),
\]
where $\hat p_t$ is the associated behavioral plan, possibly absent; $c_t$ is the selected configuration; $\rho_t$ contains the pre-evaluation rationale and advisory plan-alignment assessment; and $m_{t+1}$ collects numerical geometry and outcome information.

The full observation set $\mathcal{D}_t$ is used for numerical BO. LLM inputs instead contain a bounded projection of recent decision records and calculated context. In the main settings, at most three completed records from $\mathcal{R}_t$ are supplied to each of the three decision functions. The latest record appears once, accompanied by up to two preceding records in chronological order. Table~\ref{tab:memory_record_fields} distinguishes archived information from its prompt-visible projection.

\begin{table}[h]
    \centering
    \small
    \setlength{\tabcolsep}{5pt}
    \renewcommand{\arraystretch}{1.10}
    \caption{Stored decision-record information and its LLM-visible projection. ``All displayed records'' refers to the bounded recent-record view, not the complete archive. The latest record additionally exposes its historical intent and pre-evaluation justification.}
    \label{tab:memory_record_fields}
    \begin{tabularx}{\linewidth}{@{}p{0.18\linewidth}Xp{0.29\linewidth}@{}}
        \toprule
        Information & Archived content & LLM-visible projection \\
        \midrule
        Decision identity &
        Controller iteration and configuration, candidate, and originating-plan identifiers. &
        Controller iteration and an observation index. \\
        \addlinespace[3pt]
        Configuration &
        Selected surrogate profile, acquisition family, and search parameter values. &
        Included in all displayed records. \\
        \addlinespace[3pt]
        Behavioral intent &
        The associated historical plan, including its action, purpose, candidate preference, and revision condition, or an absent association. &
        Latest record only; plan content is retained without its local identifier. \\
        \addlinespace[3pt]
        Candidate and observation &
        The cached candidate and score. &
        Included in all displayed records. \\
        \addlinespace[3pt]
        Candidate geometry &
        Distances to the incumbent and nearest observation, together with the behavior label. &
        Included in all displayed records. \\
        \addlinespace[3pt]
        Surrounding context &
        Complete context snapshots before and after evaluation, including the incumbent. &
        Previous incumbent score in all displayed records. \\
        \addlinespace[3pt]
        Objective outcome &
        Score change from the previous incumbent as a strict improvement indicator and relative score change. &
        Included in all displayed records. \\
        \addlinespace[3pt]
        Selection feedback &
        Pre-evaluation selection rationale and advisory plan-alignment assessment. &
        Latest record only. \\
        \bottomrule
    \end{tabularx}
\end{table}

Under the maximization convention, let $b_t = \max_{(\mathbf x,y) \in \mathcal D_t} y$. The outcome fields exposed to the LLM are
\[
\Delta_{t+1} = y_{t+1} - b_t, \qquad I_{t+1} = \mathbf 1 \{ \Delta_{t+1} > 0 \}.
\]
The score change $\Delta_{t+1}$ is not scaled, and ties are not counted as improvements. Candidate geometry is computed from the selected cached point against $\mathcal D_t$ before objective evaluation. The distance convention and behavior labels are defined in Appendix~\ref{app:method_details:prospective_diagnostics}.

\paragraph{Strategy memory.}
Strategy memory $\mathcal S_t = (u_t, p_t)$ retains a revisable trajectory summary and alternative behavioral plans. The summary $u_t$ contains a cumulative evidence summary and an assessment of the latest completed decision. The behavioral-plan component retains the overall search strategy, current focus, and alternative plans. Each alternative plans specifies an action, its purpose, preferred candidate properties, and a revision condition. Planning returns a complete replacement of the reusable strategy memory, carrying forward relevant earlier interpretations within these fields. Plans are unordered alternatives for subsequent evaluation; their identifiers are local linking keys without persistent semantic meaning or priority. Plan alignment is advisory, and the final selector may associate its choice with any current plan or with none.

\paragraph{Calculated decision context.}
The common context summarizes the current observation history and completed controller decisions. Table~\ref{tab:app_decision_context_fields} specifies its fields and the history used by each calculation. All recency statistics use chronological observation order, with initialization preceding controller evaluations. In the main settings, the window size is $W = 5$, independently of the three-record display limit. Spatial statistics use the latest observations and may include initialization; behavior summaries count only completed controller decisions.

\begin{table}[h]
    \centering
    \small
    \setlength{\tabcolsep}{5pt}
    \renewcommand{\arraystretch}{1.10}
    \caption{Calculated context supplied to all three LLM decision stages. These summaries are computed from the stored history independently of the recent-record display limit.}
    \label{tab:app_decision_context_fields}
    \begin{tabularx}{\linewidth}{@{}p{0.19\linewidth}p{0.23\linewidth}X@{}}
        \toprule
        Field & History used & Definition \\
        \midrule
        Incumbent & All observations. &
        Best observed coordinates and score, with an observation index. Ties are resolved by taking the first maximizer. \\
        \addlinespace[3pt]
        Observation count & All observations. &
        Total number of evaluations, including initialization. \\
        \addlinespace[3pt]
        Stagnation & All observations in order. &
        Number of observations since the last strict incumbent improvement. \\
        \addlinespace[3pt]
        Recent spatial window & $W$. &
        Number of observations used for the recent statistics. \\
        \addlinespace[3pt]
        Recent dispersion & Recent spatial window. &
        Mean normalized distance over all pairs within the window; zero if fewer than two observations are available. \\
        \addlinespace[3pt]
        Recent novelty & Recent spatial window and nearest earlier observations. &
        Mean distance from each eligible recent observation to its nearest observation in the full history. The first observation is excluded; the statistic is zero if no comparisons exist. \\
        \addlinespace[3pt]
        Global coverage & All observations. &
        Occupied-cell count relative to expected uniform-box occupancy on a five-bin-per-coordinate grid, clipped at one. \\
        \addlinespace[3pt]
        Recent behavior outcomes & Latest $W$ controller decisions. &
        Evaluation and strict-improvement counts for each distance-based behavior label. Initialization is excluded. \\
        \addlinespace[3pt]
        Whole-run behavior & All completed controller decisions. &
        Total decisions and strict improvements, together with evaluation counts, decision fractions, and improvement counts for each behavior label. Initialization is excluded. \\
        \bottomrule
    \end{tabularx}
\end{table}

For the coverage statistic, let $n_t$ be the total number of observations, $M = 5^d$ the number of grid cells, and $K_t$ the number of occupied cells. The reported value is
\begin{equation}
\operatorname{coverage}_t = \min \left \{ 1, \, \frac{K_t}{M \left[ 1 - (1 - 1/M)^{n_t} \right]} \right\}.
\label{eq:app_grid_coverage}
\end{equation}
The denominator is the expected occupancy under independent, equally probable cell assignments, which provides a bounding-box reference for constrained spaces. A value of one indicates at least the reference occupancy.

\paragraph{Previous-generation feedback.}
Planning and shortlisting additionally receive an aggregate summary of the preceding iteration. It reports the numbers of attempted, successful, and failed distinct configurations; the number of distinct candidate points; behavior-label counts and fractions; and the minimum, median, and maximum incumbent distance and novelty across those points. It also lists the attempted surrogate profiles and acquisition parameter values, including unsuccessful BO attempts, and identifies the selected candidate's behavior label. These statistics use the preceding iteration's pre-evaluation geometry. Behavior fractions are calculated over distinct candidate points. The feedback exposes aggregate proposal availability; it does not include the previous candidate coordinates, point--configuration mappings, or outcomes for unevaluated proposals. Numerical calculations and stored candidates retain full precision. Prompt values generally use four significant digits, while domain bounds, constraints, and discrete parameter identities are preserved.

\subsection{Surrogate Fit Evidence and Configuration Shortlisting}
\label{app:method_details:fit_evidence}

\paragraph{Surrogate fit evidence.}
At each iteration, STaSBO fits every surrogate profile $g \in \mathcal G$ to the current observations $\mathcal D_t$, caches the successfully fitted models, and extracts their descriptors:
\begin{equation}
\mathcal E_t = \{(g, e_t(g)): g \in \mathcal G \}, \qquad e_t(g) = \phi_{\mathrm{fit}}(\mathcal D_t, g).
\label{eq:app_surrogate_fit_evidence}
\end{equation}
The shortlisting prompt receives the projection in Table~\ref{tab:app_fit_evidence_fields}, keyed by profile identifier and accompanied by the fixed profile specifications from Table~\ref{tab:app_surrogate_profiles}. Shared prompt instructions describe how these quantities should be interpreted.

\begin{table}[h]
    \centering
    \small
    \setlength{\tabcolsep}{5pt}
    \renewcommand{\arraystretch}{1.12}
    \caption{Surrogate fit evidence supplied to configuration shortlisting. Availability is reported separately from the fitted numerical descriptors.}
    \label{tab:app_fit_evidence_fields}
    \begin{tabularx}{\linewidth}{@{}p{0.27\linewidth}X@{}}
        \toprule
        Field & Definition and units \\
        \midrule
        Fit availability &
        Whether a usable fitted surrogate is available. Numerical descriptors are omitted for unavailable fits. \\
        \addlinespace[3pt]
        Learned length scales &
        One value per dimension for ARD profiles or one shared value for isotropic profiles. Values are expressed in the GP's training-data-normalized input coordinates. \\
        \addlinespace[3pt]
        Likelihood-noise variance &
        Learned residual-noise variance of the Gaussian likelihood, expressed in squared standardized-output units. \\
        \addlinespace[3pt]
        Kernel output scale &
        Learned multiplicative scale of the base-kernel covariance, expressed in squared standardized-output units. \\
        \addlinespace[3pt]
        Noise-to-kernel variance ratio &
        Likelihood-noise variance divided by kernel output scale. This derived quantity is dimensionless and is included when the ratio is finite and defined. \\
        \bottomrule
    \end{tabularx}
\end{table}

Input normalization uses the coordinate-wise ranges of the current training observations. Output standardization uses the observed scores, with numerical safeguards for degenerate data. The noise variance describes the fitted likelihood, and the output scale describes the kernel covariance. Length scales characterize the fitted covariance. The LLM receives an unavailable status for a failed fit, and such a profile cannot be nominated during shortlisting. If no usable surrogate is available, the iteration stops without evaluating the objective.

\paragraph{Plan-conditioned configuration shortlisting.}
The shortlisting decision combines the current strategy $\mathcal S_t$ with $\mathcal E_t$, fixed profile specifications, and the acquisition families and parameter ranges defined in Appendix~\ref{app:method_details:profile_libraries}. It also receives the shared decision context, recent records, and previous-generation feedback listed in Table~\ref{tab:app_llm_interfaces}. One LLM call proposes configurations for all current behavioral plans. For the $j$th alternative plan in $p_t$, let $\widetilde g_t^{(j)}$ denote its proposed surrogate profile and $\widetilde{\mathcal A}_t^{(j)}$ the finite set of proposed associated acquisition specifications. Each specification contains an acquisition family and an LLM-proposed parameter value: $\xi_{\mathrm{frac}}$ for EI or $\kappa$ for UCB. The plan-specific sets define the shortlist
\begin{equation}
\widetilde{\mathcal C}_t = \bigcup_{j = 1}^{|p_t|} \left \{ \bigl(\widetilde g_t^{(j)}, a \bigr): a \in \widetilde{\mathcal A}_t^{(j)} \right\}.
\label{eq:app_planwise_shortlist}
\end{equation}
The plan index is only a notational ordering of the unordered alternatives.

In the main settings, each of three plans nominates one successfully fitted surrogate and exactly three acquisition specifications. Validation requires the correct parameter for each family, a finite numerical value within its inclusive allowed range, and no identical family--parameter repetitions within a plan. Different plans may share surrogate profiles and acquisition specifications. The union merges configurations only when their surrogate identifier, acquisition family, and parameter values are identical, preserving all originating-plan associations. It therefore contains at most nine distinct configurations under the main settings. These associations record the plans that motivated each proposal. Distinct configurations remain separate at this step even if they subsequently generate identical coordinates. Candidate grouping is described in Appendix~\ref{app:method_details:prospective_diagnostics}.

\subsection{Prospective Candidate Generation and Diagnostics}
\label{app:method_details:prospective_diagnostics}

For each distinct shortlisted configuration $c\in\widetilde{\mathcal C}_t$, numerical BO generates a prospective candidate using \eqref{eq:bo_candidate}. Candidate generation reuses the surrogate fitted to $\mathcal D_t$ and the acquisition parameters specified during shortlisting. All configurations use a shared sampled pool to initialize configuration-specific continuous acquisition optimization. The optimizer searches the feasible domain defined by the task. This procedure approximates acquisition maximization without evaluating the objective. Successfully generated points are validated and cached together with their generating configurations. Configurations that produce no valid candidate are recorded as generation failures and excluded from selection. If no valid candidate remains, the iteration terminates without objective evaluation. Following Algorithm~\ref{alg:stasbo}, below $\widetilde{\mathcal C}_t$ denotes the configurations retained after this validity check.

\paragraph{Geometric diagnostics.}
For each retained configuration, the diagnostic operator computes
\begin{equation}
    z_t(c) = \phi_{\mathrm{diag}} \! \left( \mathcal D_t, c, \mathbf x_{t+1}^{(c)} \right),
    \qquad
    \mathcal Z_t = \{ (c, z_t(c)): c \in \widetilde{\mathcal C}_t \}.
\label{eq:app_prospective_diagnostics}
\end{equation}
Let $\mathbf x_t^{\mathrm{inc}}$ be the best observed point under the maximization convention, resolving ties by observation order. The incumbent and nearest-observation distances are
\begin{equation}
    d_t^{\mathrm{inc}}(c) = d_{\mathrm{norm}} \! \left( \mathbf x_{t+1}^{(c)},\mathbf x_t^{\mathrm{inc}} \right), 
    \qquad
    d_t^{\mathrm{near}}(c) = \min_{(\mathbf x_i, y_i) \in \mathcal D_t} d_{\mathrm{norm}} \! \left( \mathbf x_{t+1}^{(c)}, \mathbf x_i \right).
\label{eq:app_candidate_distances}
\end{equation}
The former measures displacement from the incumbent, while the latter measures geometric novelty relative to previously evaluated points, including initialization.

The normalized distance is the root-mean-square coordinate difference:
\begin{equation}
    d_{\mathrm{norm}}(\mathbf x, \mathbf x') = \sqrt{ \frac{1}{d} \sum_{j = 1}^{d} \delta_j(\mathbf x, \mathbf x')^2}.
\label{eq:app_normalized_distance}
\end{equation}
For the numerical coordinates of the synthetic and application benchmarks,
\[
\delta_j(\mathbf x, \mathbf x') = \frac{|x_j - x'_j|}{u_j - \ell_j},
\]
where $[\ell_j, u_j]$ is the declared parameter range. The general parameter-space metric uses logarithmic range differences for logarithmic parameters and a zero/one mismatch for categorical parameters.

The incumbent distance determines the behavior label:
\[
\mathrm{label}_t(c) = 
\begin{cases}
    \text{local refinement}, & d_t^{\mathrm{inc}}(c) \leq 0.05,\\
    \text{intermediate move}, & 0.05<d_t^{\mathrm{inc}}(c) \leq 0.15,\\
    \text{broad move}, & d_t^{\mathrm{inc}}(c) > 0.15.
\end{cases}
\]
The same convention is used for decision-record geometry. These labels describe spatial displacement; they do not measure posterior uncertainty or establish candidate quality.

\paragraph{Candidate grouping and presentation.}
After generation, candidates are grouped when their validated, full-precision encoded coordinates are exactly equal. This operation is separate from configuration deduplication during shortlisting. Nearby points remain in different groups, and each group retains every configuration that generated its point.

Table~\ref{tab:app_candidate_diagnostic_fields} summarizes the information supplied to the selector. Display rounding occurs after grouping and does not modify cached points. Distinct candidates can therefore appear numerically identical in the prompt while remaining separate selectable actions. Candidate-generation failures are supplied separately with the associated configuration and error information.

\begin{table}[h]
    \centering
    \small
    \setlength{\tabcolsep}{5pt}
    \renewcommand{\arraystretch}{1.12}
    \caption{Candidate-group information supplied to joint configuration selection. Each group corresponds to one exact cached point and retains all of its generating configurations.}
    \label{tab:app_candidate_diagnostic_fields}
    \begin{tabularx}{\linewidth}{@{}p{0.27\linewidth}X@{}}
        \toprule
        Field & Content \\
        \midrule
        Candidate identifier &
        An iteration-local identifier for the exact-point group. \\
        \addlinespace[3pt]
        Candidate coordinates &
        The validated point in the task's original parameter units. \\
        \addlinespace[3pt]
        Incumbent distance &
        Normalized distance to the current best observed point, $d_t^{\mathrm{inc}}(c)$. \\
        \addlinespace[3pt]
        Nearest-observation distance &
        Minimum normalized distance to any previously evaluated point, $d_t^{\mathrm{near}}(c)$. \\
        \addlinespace[3pt]
        Behavior label &
        Local refinement, intermediate move, or broad move, determined by incumbent distance. \\
        \addlinespace[3pt]
        Generating configurations &
        Configuration identifiers, surrogate profiles, acquisition families and parameter values, and originating-plan identifiers. \\
        \bottomrule
    \end{tabularx}
\end{table}

\subsection{Joint Configuration Selection and Feedback}
\label{app:method_details:joint_selection}

The final LLM decision selects one retained configuration:
\begin{equation}
(c_t, \hat p_t, \rho_t) = \pi_{\mathrm{select}} \! \left( \mathcal S_t, \widetilde{\mathcal C}_t, \mathcal Z_t; \mathcal P \right),
\qquad
c_t \in \widetilde{\mathcal C}_t.
\label{eq:app_joint_selection}
\end{equation}
As detailed in Table~\ref{tab:app_llm_interfaces}, the selector also receives the calculated current context, bounded recent decision records, static surrogate definitions, and generation-failure information. Selection therefore compares the available candidate behaviors using the strategy and evaluated search history.

\paragraph{Configuration choice and plan association.} 
The selector is instructed to examine the available candidates against each unordered behavioral plan, identify supported or partial matches, and compare their usefulness for the next evaluation. It returns one successfully generated configuration. The implementation resolves the corresponding candidate identifier and cached point from this configuration identifier. When several configurations generate the same point, the output still identifies one actual generating configuration for the decision record. The associated plan $\hat p_t$ may refer to any current behavioral plan or be absent. Plan identifiers record why configurations were proposed; they do not restrict the final association. Partially aligned and outside-plan candidates remain admissible. The pre-evaluation feedback $\rho_t$ contains a concise decision rationale and an advisory plan-alignment assessment. The assessment lists relevant requirements, a status of \texttt{met}, \texttt{not\_met}, or \texttt{uncertain} for each, supporting evidence, and limitations. The rationale explains the selected action's intended contribution, relevant alternatives, and uncertainty.

\paragraph{Evaluation and feedback update.}
The selected point $\mathbf x_{t+1}^{(c_t)}$ is passed directly to the objective evaluator. Before evaluation, its geometry is computed against $\mathcal D_t$. The returned coordinates and score form the observation $o_{t+1}=(\mathbf x_{t+1},y_{t+1})$. The observation and pre-evaluation context determine the numerical record fields $m_{t+1}$, including the score change relative to the previous incumbent and the strict improvement indicator defined in Appendix~\ref{app:method_details:memory_details}. The associated plan, selected configuration, and pre-evaluation feedback are retained in
\[
r_{t+1}=(\hat p_t,c_t,\rho_t,m_{t+1}),
\]
and history memory is updated through
\[
\mathcal D_{t+1} = \mathcal D_t\cup\{o_{t+1}\}, \qquad \mathcal R_{t+1} = \mathcal R_t\cup\{r_{t+1}\}.
\]
The generation diagnostics, failures, and selected configuration are also retained to construct the next iteration's aggregate generation feedback. Unselected candidates retain their prospective diagnostics but receive no measured objective outcomes. At iteration $t+1$, planning uses the updated history and feedback to revise the trajectory summary and behavioral plans.

\lstdefinestyle{stasboPrompt}{
    basicstyle=\footnotesize\ttfamily,
    columns=fullflexible,
    keepspaces=true,
    breaklines=true,
    breakatwhitespace=false,
    showstringspaces=false,
    frame=single,
    framesep=4pt,
    captionpos=t,
    aboveskip=8pt,
    belowskip=8pt
}

\lstdefinestyle{stasboDescription}{
    style=stasboPrompt,
    basicstyle=\small\rmfamily,
    breakatwhitespace=true
}

\section{Experimental Details}
\label{app:implementation_details}

This section describes the benchmarks, evaluation protocol, baseline implementations, and computational settings used in the main experiments. The sensitivity experiments in Section~2 use the separate protocol described in Appendix~\ref{app:sensitivity_analysis}. STaSBO's controller interfaces, memory contents, and evidence definitions are specified in Appendix~\ref{app:method_details}.

\subsection{Benchmark Definitions}
\label{app:implementation_details:benchmark_definitions}

We evaluate 12 synthetic mathematical tasks and 10 scientific and engineering application tasks. The synthetic suite combines four functions with three dimensionalities. The application suite contains nine objectives constructed from experimental datasets and one physics-based simulation.

\paragraph{Synthetic mathematical benchmarks.}
We evaluate Ackley, Levy, Rastrigin, and Rosenbrock in $d\in\{5,8,15\}$. Table~\ref{tab:app_main_synthetic_domains} specifies their minimization forms, search domains, and random shift distributions. For each task and seed, we independently sample the coordinates of $\boldsymbol{\delta}$ and evaluate the maximization score $y(\mathbf{x}) = -h(\mathbf{x} - \boldsymbol{\delta})$. The translated minimizer lies within the central $80\%$ of each coordinate range, while the search bounds remain unchanged. All four functions have a minimum value of zero, and no observation noise is added.

\begin{table}[h]
    \centering
    \caption{Synthetic mathematical benchmarks. Each function is evaluated in 5D, 8D, and 15D. Translation coordinates are drawn independently from the listed uniform distributions.}
    \label{tab:app_main_synthetic_domains}
    \footnotesize
    \setlength{\tabcolsep}{3pt}
    \renewcommand{\arraystretch}{1.25}
    \begin{tabularx}{\linewidth}{@{}lXcc@{}}
        \toprule
        Function & Objective $h(\mathbf z)$ & Search domain & Shift $\delta_j$ \\
        \midrule
        
        Ackley &
        $\begin{aligned}[t]
        &-20\exp\!\left(-0.2\sqrt{\frac{1}{d}\sum_{j=1}^{d}z_j^2} \right)\\
        &-\exp\!\left( \frac{1}{d}\sum_{j=1}^{d}\cos(2\pi z_j) \right) + 20 + e
        \end{aligned}$
        & $[-32.768,32.768]^d$ & $\mathcal U[-26.2144,26.2144]$ \\[5pt]
        
        Levy &
        $\begin{aligned}[t]
        &\sin^2(\pi w_1)+\sum_{j=1}^{d-1}(w_j-1)^2 \bigl[1+10\sin^2(\pi w_j+1)\bigr]\\
        &+(w_d-1)^2 \bigl[1+\sin^2(2\pi w_d)\bigr]
        \end{aligned}$
        & $[-10,10]^d$ & $\mathcal U[-9,7]$ \\[5pt]
        
        Rastrigin & $\displaystyle 10d + \sum_{j=1}^{d} \bigl[z_j^2 - 10 \cos(2 \pi z_j)\bigr]$ & $[-5.12,5.12]^d$ & $\mathcal U[-4.096,4.096]$ \\[5pt]
        
        Rosenbrock & $\displaystyle \sum_{j=1}^{d-1} \Bigl[ 100(z_{j+1} - z_j^2)^2 + (1 - z_j)^2\Bigr]$ & $[-2.048,2.048]^d$ & $\mathcal U[-2.6384,0.6384]$ \\
        
        \bottomrule
    \end{tabularx}

    \smallskip
    \begin{minipage}{\linewidth}
    \footnotesize
    Here, $\mathbf z=\mathbf x-\boldsymbol{\delta}$ and $w_j=1+(z_j-1)/4$ for Levy. The unshifted minimizers are $\mathbf z=\mathbf 0$ for Ackley and Rastrigin and $\mathbf z=\mathbf 1$ for Levy and Rosenbrock. The internal maximization score is $-h(\mathbf z)$.
    \end{minipage}
\end{table}

\paragraph{Dataset-based application benchmarks.}
Table~\ref{tab:app_application_benchmarks} summarizes the application tasks. Concrete uses the UCI Concrete Compressive Strength dataset \citep{concrete_compressive_strength_165}. The remaining experimental datasets are distributed through Olympus~\citep{hase2021olympus, hickman2023olympus}. We construct numerical objectives from the archived measurements using our own interpolation procedure. The resulting objectives are fixed throughout optimization.

\begin{table}[h]
    \centering
    \caption{Scientific and engineering application benchmarks. $N$ is the number of distinct design points after preprocessing; $d$ counts represented input coordinates. Arrows indicate the implemented optimization direction. Each OER task has six composition coordinates and five independent degrees of freedom.}
    \label{tab:app_application_benchmarks}
    \small
    \setlength{\tabcolsep}{4pt}
    \renewcommand{\arraystretch}{1.12}
    \begin{tabularx}{\linewidth}{@{}l>{\raggedright\arraybackslash}p{0.3\linewidth}rr>{\raggedright\arraybackslash}X@{}}
        \toprule
        Task & Domain & $d$ & $N$ & Objective \\
        \midrule
        Concrete & Civil engineering & 7 & 427 & Compressive strength $\uparrow$ \\ 
        Crossed Barrel & Mechanical design & 4 & 600 & Toughness $\uparrow$ \\
        AgNP & Nanomaterials synthesis & 5 & 164 & Archived spectrum score $\uparrow$ \\
        P3HT & Functional materials & 5 & 178 & Electrical conductivity $\uparrow$ \\
        HPLC & Analytical chemistry & 6 & 1,007 & Recorded peak area $\uparrow$ \\
        OER 3496 & Electrocatalysis & 6 & 2,121 & Overpotential $\downarrow$ \\ 
        OER 3851 & Electrocatalysis & 6 & 2,119 & Overpotential $\downarrow$ \\
        OER 3860 & Electrocatalysis & 6 & 2,120 & Overpotential $\downarrow$ \\
        OER 4098 & Electrocatalysis & 6 & 2,121 & Overpotential $\downarrow$ \\
        Lunar Lander & Control & 12 & --- & Mean episode return $\uparrow$ \\
        \bottomrule
    \end{tabularx}
\end{table}

Rows with missing values are removed, and measurements sharing identical design coordinates are averaged. For Concrete, design coordinates comprise seven mixture ingredients and exclude curing age. Consequently, measurements of the same mixture can be averaged across different ages. For AgNP, the implementation follows the maximization direction declared by Olympus. However, the archived target values originate from the spectral loss minimized in the original experiment and are not sign-reversed. Higher values therefore indicate better performance under the implemented benchmark convention, but should not be interpreted as improved physical spectral matching. For HPLC, the objective is recorded peak area, rather than a combined measure of separation quality, analysis time, or solvent consumption.

\begin{table}[h]
    \centering
    \small
    \caption{Application design variables and numerical search domains. Bounds are independent except for the OER simplex constraints. All represented variables are continuous.}
    \label{tab:app_application_domains}
    \begin{tabularx}{\linewidth}{@{}p{0.17\linewidth}X@{}}
        \toprule
        Task & Variables, bounds, and constraints \\
        \midrule
        Concrete & Cement $[102, 540]$, blast-furnace slag $[0, 359.4]$, fly ash $[0, 200.1]$, water $[121.8, 247]$, superplasticizer $[0,32.2]$, coarse aggregate $[801,1145]$, and fine aggregate $[594,992.6]$, in $\mathrm{kg}/\mathrm{m}^3$. No fixed-total constraint is imposed. \\
        Crossed Barrel & Column count $[6,12]$, twist angle $[0,200]$ degrees, outer radius $[1.5,2.5]$ mm, and wall thickness $[0.7,1.4]$ mm. Column count is continuously relaxed. \\
        AgNP & Feed ratios for silver nitrate $[4.53, 42.80981595]$, polyvinyl alcohol $[9.999518096, 40.00101474]$, trisodium citrate $[0.5, 30.5]$, and silver seeds $[0.498851653, 19.5]$, on the recorded percentage scale; total flow $[200, 983] \, \mu \mathrm{L} /\mathrm{min}$. No additional coupled flow constraint is imposed. \\
        P3HT & P3HT $[15, 96.27]$, long single-walled nanotubes $[0,60]$, short single-walled nanotubes $[0, 70]$, multi-walled nanotubes $[0, 85]$, and double-walled nanotubes $[0, 75]$, on the recorded percentage scale. No fixed-sum constraint is imposed. \\
        HPLC & Sample-loop volume $[0, 0.08]$ mL, additional volume $[0, 0.06]$ mL, tubing volume $[0.1, 0.9]$ mL, sample flow $[0.5,2.5]$ mL/min, push speed $[80,150]$ Hz, and wait time $[0.5,10]$ s. \\
        OER 3496 & Relative loadings of Ni, Fe, Co, Mn, Ce, and La. \\
        OER 3851 & Relative loadings of Ni, Fe, Co, Ta, Mn, and Cu. \\
        OER 3860 & Relative loadings of Sn, Fe, Co, Ta, Mn, and Cu. \\
        OER 4098 & Relative loadings of Sn, Sb, Co, Ca, Ni, and Mn. \\
        Lunar Lander & Twelve feedback-controller coefficients, each in $[0,2]$. \\
        \bottomrule
    \end{tabularx}
    
    \vspace{3pt}
    \begin{minipage}{\linewidth}
        \footnotesize
        For every OER task, $x_j \in [0, 1]$ and $\sum_{j = 1}^{6}x_j = 1$. P3HT measurements have approximately fixed-total compositions, but the implemented search domain is an unconstrained box.
    \end{minipage}
\end{table}

\paragraph{Objective interpolation.}
All nine processed experimental datasets are irregularly sampled, and their numerical objectives use inverse-distance weighting over the $k = 12$ nearest measured designs. For a query $\mathbf{x}$, distances are computed after scaling each coordinate by its declared domain width:
\[
    d_\ell(\mathbf{x}) = \sqrt{ \sum_{j = 1}^{d} \left( \frac{x_j - x_{\ell, j}}{u_j - l_j} \right)^2},
\]
where $l_j$ and $u_j$ denote the coordinate bounds. Given neighboring measurements $v_\ell$, the interpolated value is
\begin{equation}
    \widehat v(\mathbf{x}) =
    \frac{
        \sum_{\ell \in \mathcal N_{12}(\mathbf{x})}
        (d_\ell(\mathbf{x}) + \varepsilon)^{-2}v_\ell
    }{
        \sum_{\ell \in \mathcal N_{12}(\mathbf{x})}
        (d_\ell(\mathbf{x}) + \varepsilon)^{-2}
    },
    \qquad \varepsilon = 10^{-8}.
\label{eq:app_objective_interpolation}
\end{equation}
If a neighbor lies within $\varepsilon$, the evaluator instead returns the mean measurement among the matching neighbors. These evaluations are deterministic, with no additional observation noise. Measurements to be minimized are negated before entering the optimizer, so all internal scores follow a maximization convention. The reference for each dataset-based task is its best processed dataset score. This reference describes the numerical benchmark and is not asserted to be the global optimum of the underlying physical system.

\paragraph{Lunar Lander.}
The Lunar Lander benchmark optimizes the twelve coefficients of a TuRBO-style heuristic feedback controller in Gymnasium's \texttt{LunarLander-v3} environment. The coefficients determine a policy over four discrete actions, while the optimization domain is the continuous box $[0, 2]^{12}$. Each objective evaluation averages the cumulative return over 50 episodes with the fixed episode seeds $0,\ldots,49$. The same episode set is used for every candidate and optimization seed. Each episode is limited to 1,000 steps; gravity is $-10$, wind is disabled, and the initial random-force scale is 1,500. The recorded runs use Gymnasium 1.4.0 and Box2D 2.3.10. Thus, one objective evaluation means evaluating one controller over the complete episode set, rather than running one episode.

We use a fixed reference return of 300 for reporting. This value is neither a known global optimum nor an upper bound, so the reported quantity is a reference gap. Episode-to-episode variation is distinct from repeated-evaluation noise: under the fixed simulator and episode seeds, repeated evaluations of the same coefficients reproduce the same mean return.

\subsection{Initialization and Evaluation Protocol}
\label{app:implementation_details:evaluation_protocol}

\paragraph{Evaluation budget and initialization.}
Each run starts with $n_0 = 5$ observations and performs $T = 45$ additional sequential evaluations, for a total budget of 50 objective evaluations. Every optimization iteration evaluates one candidate. All experiments use ten seeds, $s\in\{0,\ldots,9\}$. Methods share the initial observations for each task and seed. Box-constrained domains use Latin hypercube sampling. OER domains use $\operatorname{Dirichlet}(\mathbf{1})$ sampling to obtain feasible simplex compositions. Iteration~0 in trajectory plots denotes the incumbent immediately after these five initial evaluations.

\paragraph{Regret and reference gap.}
Let $y_{m,j,s,i}$ be the maximimum score obtained by method $m$ on task $j$, seed $s$, at iteration $i$. The incumbent and reporting gap after $t$ additional evaluations are
\begin{equation}
    \begin{aligned}
        b_{m,j,s}(t) &= \max_{1\leq i\leq n_0+t} y_{m,j,s,i},
        \qquad
        r_{m,j,s}(t) &= y^{\mathrm{ref}}_{j,s}-b_{m,j,s}(t).
    \end{aligned}
    \label{eq:app_reporting_gap}
\end{equation}
For synthetic tasks, $y^{\mathrm{ref}}_{j,s}=0$ is the known optimal score, and $r_{m,j,s}$ is simple regret. For dataset-based applications, the reference is the best score in the dataset. For Lunar Lander, it is the fixed return of 300.

\paragraph{Task-level and average ranks.}
We first average final regrets over seeds:
\begin{equation}
    \overline r_{m, j}(T) = \frac{1}{10} \sum_{s = 0}^{9}r_{m, j, s}(T).
\label{eq:app_seed_mean_gap}
\end{equation}
Methods are then ranked within each task using these means, with lower values receiving better ranks. Exact ties receive average ranks. The average-rank column is the arithmetic mean of task-level ranks within the corresponding benchmark suite.

\paragraph{Random-normalized aggregate trajectories.}
Raw regrets can have different units and scales across tasks. We normalize each method's seed-averaged trajectory by baseline Random's seed-averaged final regret on the same task:
\begin{equation}
    q_{m, j}(t) = \frac{\frac{1}{10}\sum_{s = 0}^{9}r_{m, j, s}(t)}{\frac{1}{10}\sum_{s = 0}^{9} r_{\mathrm{Random}, j, s}(T)}.
\label{eq:app_random_normalization}
\end{equation}
The denominator is fixed across iterations. Accordingly, Random ends at one, while its earlier values need not equal one.

For a benchmark suite $\mathcal T$, the aggregate trajectory is
\begin{equation}
    Q_{m, \mathcal T}(t) = \frac{1}{|\mathcal T|} \sum_{j \in \mathcal T}q_{m, j}(t).
\label{eq:app_suite_aggregation}
\end{equation}
In Figure~\ref{fig:aggregated_results}(a), $\mathcal T$ contains all 22 benchmark tasks. Task-level normalized final results report $q_{m,j}(T)$.

\paragraph{Iterations to a target regret.}
For a normalized-regret target $\gamma$, we identify the first observed BO iteration at which the aggregate curve reaches the target:
\[
\tau_m(\gamma) = \min \left \{ t \in \{0, \ldots, T \}: Q_{m, \mathcal T}(t) \leq \gamma \right \}.
\]
Comparisons use the earliest-reaching baseline at each target. The relative reduction in BO iterations is $(\tau_{\mathrm{baseline}} - \tau_{\mathrm{STaSBO}}) / \tau_{\mathrm{baseline}}$. They measure threshold attainment by the aggregate mean curve, rather than the average attainment time of individual runs.

\paragraph{Geometric mean rarity.}
Rarity measures incumbent quality relative to uniform sampling over each task's feasible domain. Let $\nu_j$ denote the uniform distribution over the search box for task $j$, or over the feasible simplex for an OER task. For method $m$, task $j$, and seed $s$, let $b_{m, j, s}(t)$ be the best observed maximization score after initialization and $t$ additional BO evaluations. We define the reference tail probability and estimated incumbent rarity as
\begin{equation}
    p_{m, j, s}(t) = \Pr_{\mathbf x \sim \nu_j} \! \left[f_{j, s}(\mathbf x) \ge b_{m, j, s}(t)\right],
    \qquad
    N_{m, j, s}(t) = \frac{1}{\widehat p_{m, j, s}(t)}.
\label{eq:incumbent_rarity}
\end{equation}
The event includes points that match or exceed the incumbent's objective value. Larger rarity indicates a smaller fraction of the feasible domain attaining that quality; for example, $N=10^4$ corresponds to a tail probability of $0.01\%$.

\paragraph{Aggregation across seeds and tasks.}
Figure~\ref{fig:aggregated_results}(b) reports the geometric mean of incumbent rarity over the twelve synthetic tasks and nine dataset-based application tasks. Let $\mathcal T_{\mathrm{rar}}$ denote this set of 21 tasks. Lunar Lander is excluded because no reference tail distribution is estimated for its simulator-based objective. At each iteration, we compute rarity separately for each run, average its logarithm over ten seeds within each task and then equally across tasks, and exponentiate:
\begin{equation}
    G_m(t) = \exp \! \left[\frac{1}{|\mathcal T_{\mathrm{rar}}|}\sum_{j \in \mathcal T_{\mathrm{rar}}} \frac{1}{10}\sum_{s = 0}^{9}\log N_{m, j, s}(t) \right].
\label{eq:geometric_mean_rarity}
\end{equation}
Thus, each task receives equal weight, as does each seed within a task. The figure uses a logarithmic vertical axis; higher values indicate better aggregate performance under the reference distributions.

\paragraph{Reference estimation.}
Tail probabilities are estimated after optimization using the same analytic functions and fixed application interpolants that define the benchmarks. Ackley and Rosenbrock use enclosing-region or change-of-variables estimators with known volume factors, with direct uniform sampling where applicable. Levy and Rastrigin use importance sampling with evaluated proposal densities. Application tails use defensive-mixture importance sampling over their continuous boxes or simplices. These estimators preserve the uniform reference measure, and the same reference mapping is used across methods for each benchmark instance. All included tail-probability estimates are positive and finite, with estimated relative Monte Carlo standard errors at most $10\%$. No probability floor or rarity cap is applied in the aggregation. Reference evaluations are used only for post-hoc assessment and are never supplied to the optimization methods.

\subsection{Baseline Implementations}
\label{app:implementation_details:baselines}

All methods share the tasks, seeded initial observations, objective-evaluation budget, and reporting protocol. Baseline-specific surrogate models, acquisition definitions, and numerical procedures are retained. Table~\ref{tab:app_baseline_settings} summarizes the principal settings.

\begin{table}[h]
    \centering
    \caption{Principal baseline settings in the main comparison. Here, $d$ is the input dimension and $n$ is the number of accumulated observations. Gamma distributions use shape--rate parameterization.}
    \label{tab:app_baseline_settings}
    \small
    \renewcommand{\arraystretch}{1.12}
    \begin{tabularx}{\linewidth}{@{}p{0.17\linewidth}X@{}}
        \toprule
        Method & Configuration and settings \\
        \midrule
        
        Random &
        Latin hypercube sampling for boxes; $\operatorname{Dirichlet}(\mathbf 1)$ sampling for OER simplices. \\
        
        BO-EI &
        Mat\'ern-$5/2$ ARD GP; EI with $\xi_{\mathrm{frac}}=0$. \\
        
        BO-UCB &
        Mat\'ern-$5/2$ ARD GP; UCB with $\kappa=3$. \\
        
        GP-Hedge &
        EI/PI/UCB portfolio; softmax parameter $\eta=1$. \\
        
        No-PASt-BO &
        EI/PI/UCB portfolio; memory coefficient $0.7$ and $\eta=4$. \\
        
        SETUP-BO &
        EI/PI/UCB portfolio; initial memory and softmax-parameter distributions $\operatorname{Beta}(17,3)$ and $\operatorname{Gamma}(40,10)$. \\
        
        TuRBO &
        TuRBO-1 with Thompson sampling; $\min(100d,5000)$ candidates. Initial/minimum/maximum trust-region lengths: $0.8/2^{-7}/1.6$. Success/failure tolerances: $3/\max(4,d)$. GP fitting: 50 Adam steps, learning rate $0.1$. \\
        
        BOOST &
        Four isotropic kernels $\times$ four acquisitions. Reference size $\min\{20,\max\{3,\lfloor n/3\rfloor\}\}$; at most 20 replay steps per pair; UCB $\kappa=0.1$. GP fitting: 50 Adam steps, learning rate $0.05$. \\
        
        Direct LLM &
        One direct candidate-proposal call per iteration; complete observation history; random feasible fallback for invalid responses. \\
        
        LMABO &
        Fixed Mat\'ern-$5/2$ ARD GP; twelve acquisition options; one acquisition-selection call per iteration. \\
        
        CAKE &
        Six retained kernels; one crossover per iteration; mutation probability $0.7$; at most twelve terms per kernel expression. \\
        
        \bottomrule
    \end{tabularx}
\end{table}

\paragraph{Random and fixed configurations.}
Random shares the five initialization observations with the other methods, then evaluates a separately sampled sequence of 45 designs. BO-EI and BO-UCB retain their surrogate family and acquisition parameters while refitting GP hyperparameters to accumulated observations. Both use a scaled Mat\'ern-$5/2$ ARD kernel with learned length scales, output scale, and observation noise. Their numerical implementation follows Appendix~\ref{app:implementation_details:numerical_bo_implementation}.

\paragraph{Acquisition portfolios.}
GP-Hedge~\citep{hoffman2011portfolio}, No-PASt-BO~\citep{vasconcelos2019no}, and SETUP-BO~\citep{vasconcelos2022self} share an EI/PI/UCB portfolio and a scaled Mat\'ern-$5/2$ ARD GP. Inputs use declared-bound normalization, outputs are standardized, and noise variance is fixed at $10^{-6}$ in standardized units. After each evaluation, posterior means from the refitted GP at all three proposed points provide rewards. GP-Hedge accumulates rewards; No-PASt-BO discounts gains and normalizes their range before selection. SETUP-BO additionally updates and samples its memory coefficient and softmax parameter using incumbent improvement and the selected normalized gain gap.

\paragraph{TuRBO.}
We implement sequential TuRBO-1 with Thompson sampling~\citep{eriksson2019scalable}. A Mat\'ern-$5/2$ ARD GP is fitted to observations collected since the current restart, and a length-scale-weighted trust region expands or contracts according to consecutive outcomes. Box-domain candidates use scrambled Sobol sampling, followed by selection through a joint observation-posterior sample. For OER tasks, 64-step parallel hit-and-run chains generate candidates within the intersection of the simplex and trust region. Restart initialization consumes the remaining evaluation budget.

\paragraph{BOOST.}
BOOST~\citep{park2025boost} compares isotropic Mat\'ern-$3/2$, Mat\'ern-$5/2$, RBF, and rational-quadratic kernels paired with EI, PI, UCB, or posterior mean. It constructs a reference set by clustering observations whose losses exceed a target order statistic near the fifth percentile. Each pair replays BO over the remaining recorded observations, reusing their known outcomes. The pair with the lowest replay score is fitted to all observations and generates the next query. Replay requires no additional objective evaluations. Its GPs use training-subset input normalization, median-centered and scaled losses, and learned observation noise.

\paragraph{Direct LLM.}
This baseline directly proposes one evaluation point from the problem description, declared search space and constraints, complete observation history, incumbent, and remaining budget. It uses no GP or acquisition function and carries no generated summaries or rationales between iterations. Each iteration permits one generation attempt. Invalid or incomplete responses trigger a logged random feasible proposal without a repair call; API and objective-evaluation failures propagate. Valid duplicate proposals are evaluated and count toward the budget.

\paragraph{LMABO.}
LMABO~\citep{ngo2026adaptive} retains a scaled Mat\'ern-$5/2$ ARD GP with learned noise and selects among PI, LogPI, EI, LogEI, UCB, posterior mean, posterior standard deviation, Thompson sampling, qKG, qPES, qMES, and qJES. It makes one initial acknowledgement request and one selection request per iteration, retaining the conversation history. Malformed selections fall back to UCB. It uses native, acquisition-specific optimization and sampling procedures; their internal samples do not consume objective evaluations.

\paragraph{CAKE.}
CAKE~\citep{suwandi2026adaptive} initializes squared-exponential, periodic, linear, rational-quadratic, Mat\'ern-$3/2$, and Mat\'ern-$5/2$ kernels with ARD parameterizations. LLM crossover and mutation generate composite kernels through sums and products, and the six lowest-BIC kernels are retained. Each kernel proposes an EI-maximizing candidate. Final query selection uses BIC-weighted EI, computed as $\log w_k+\operatorname{LogEI}_k$, where $w_k\propto\exp(-\mathrm{BIC}_k)$. Malformed crossover outputs use random composition, while malformed mutations are skipped.

Acquisition-optimization budgets for the shared numerical proposer are reported in Appendix~\ref{app:implementation_details:numerical_bo_implementation}. TuRBO instead uses its own trust-region candidate generator and Thompson-sampling procedure.

\subsection{Controller and LLM Settings}
\label{app:implementation_details:controller_implementation}

\paragraph{Problem descriptions.}
The main experiments use the configured task descriptions. These specify objective semantics, parameter meanings, search-space constraints, and qualitative landscape or domain priors. Some descriptions also discuss modeling assumptions or search uncertainty. Synthetic descriptions disclose the optimal score of zero but do not reveal the translated optimizer location or shift vector. For STaSBO, task-agnostic control refers to using the same controller, profile space, and stage templates across tasks; the controller can still condition its decisions on supplied task context. Listings~\ref{lst:app_description_ackley} and~\ref{lst:app_description_p3ht} represent two descriptions used in the experiments. The remaining descriptions and complete prompt templates will be provided as supplementary.

\begin{lstlisting}[style=stasboDescription, caption={Problem description supplied for Ackley.}, label={lst:app_description_ackley}]
The score is maximized and has a global maximum of 0 at an undisclosed location, which may be shifted away from the coordinate origin or domain center. A distance-dependent component rises toward a pronounced peak and becomes comparatively flat far from it, while coordinate-wise oscillations produce additional local optima. The resulting landscape is continuous and smooth away from the global optimum, with a nonsmooth peak at that optimum. Local score differences reflect both the broad distance trend and the oscillations; they do not by themselves establish a coordinate-wise direction toward the optimum or a preferred step size.
\end{lstlisting}

\begin{lstlisting}[ style=stasboDescription, caption={Problem description supplied for P3HT.}, label={lst:app_description_p3ht}]
Objective and experimental context:
This benchmark concerns drop-cast P3HT/carbon-nanotube composite blends. The target column is conductivity. The supplied y = conductivity is maximized, so higher values are better. This is a conductivity objective, not photovoltaic efficiency or a combined material-performance score.

Design parameters (the exact dictionary keys are listed below):
- p3ht_content: recorded P3HT component content, in [15, 96.27].
- d1_content: recorded d1 component content, in [0, 60].
- d2_content: recorded d2 component content, in [0, 70].
- d6_content: recorded d6 component content, in [0, 85].
- d8_content: recorded d8 component content, in [0, 75].

Search-space representation:
The five continuous inputs use the recorded percentage-scale values, not fractions from 0 to 1. Although measured compositions sum approximately to 100, Olympus declares independent bounds with no simplex constraint for this benchmark. Do not impose a fixed sum or normalize proposed inputs. The numerical benchmark consequently admits settings outside the measured composition surface; those settings need not represent physically feasible blends. The component labels do not establish a universally favorable substitution, preferred region, or particular function smoothness.

Benchmark representation:
The numerical objective uses distance-weighted interpolation of archived measurements, not a new materials experiment or Olympus's pretrained emulator. Identical parameter rows are grouped and their conductivity values averaged before interpolation. Sparse observations leave uncertainty about unmeasured component balances; neither nearby nor distant proposals are guaranteed to improve the score. No optimum or optimal blend is supplied.
\end{lstlisting}

\paragraph{STaSBO settings.}
STaSBO uses the seven surrogate profiles and continuously parameterized EI/UCB acquisition space defined in Appendix~\ref{app:method_details:profile_libraries}. Planning maintains three alternative behavioral plans. For each plan, shortlisting proposes one surrogate profile and three acquisition specifications. Their union contains at most nine distinct configurations after exact configuration deduplication.

Table~\ref{tab:app_stasbo_settings} gives the detailed settings. The recent-statistics window and displayed-record limit are separate parameters: the former determines summary calculations, while the latter limits the decision records included in prompts. Raw observation arrays are not directly included in the three LLM inputs; the numerical components retain access to all observations. The actual context fields and stage-specific inputs are described in Appendix~\ref{app:method_details:memory_details}.

\begin{table}[h]
    \centering
    \small
    \caption{STaSBO controller settings used in the main experiments.}
    \label{tab:app_stasbo_settings}
    \begin{tabularx}{\linewidth}{@{}p{0.52\linewidth}X@{}}
        \toprule
        Parameter & Setting \\
        \midrule
        Surrogate profiles & 7 \\
        Behavioral plans & 3 \\
        Acquisition specifications per plan & 3 \\
        EI improvement-margin fraction & $\xi_{\mathrm{frac}}\in[0,1]$ \\
        UCB coefficient & $\kappa\in[0,20]$ \\
        Recent-statistics window & 5 \\
        Maximum displayed decision records & 3 \\
        Prompt numerical precision & Generally 4 significant digits \\
        Maximum response attempts per decision & 5, including the first attempt \\
        \bottomrule
    \end{tabularx}
\end{table}

Strategy memory begins with neutral placeholders and is updated by the first planning call after initialization. LLM outputs are validated against their required schemas and admissible choices. Invalid outputs are retried with validation feedback, subject to the response-attempt limit. Numerical values used for fitting, acquisition optimization, and objective evaluation retain full precision.

\subsection{Controller Query Templates and Response Formats}
\label{app:implementation_details:prompt_formats}

Each stage combines a fixed system prompt with a query template populated by JSON-serialized inputs. The stage responsibilities and input semantics are described in Appendices~\ref{app:method_details:overall_algorithm} and~\ref{app:method_details:memory_details}. The query templates below preserve the implemented section headings and substitution fields. Braced placeholders are replaced with the corresponding JSON objects at runtime. Response templates retain all requested fields, with shortened explanatory strings and one representative item per list. Their displayed list lengths are schematic: the runtime contracts specify the required plan identifiers and acquisition counts. These templates are illustrative JSON structures rather than formal JSON Schema documents.

\paragraph{Strategy planning.}
The planning query combines current context and recent records with previous strategy memory and aggregate feedback from the preceding candidate-generation step. The planning contract contains the required plan identifiers, \texttt{A}, \texttt{B}, and \texttt{C}.

\begin{lstlisting}[style=stasboPrompt, caption={Strategy-planning query template.}, label={lst:app_planning_query}]
[ENVIRONMENT]
{environment_json}

[CURRENT CONTEXT]
{decision_context_json}

[RECENT RECORDS]
{recent_records_json}

[PREVIOUS ITERATION FEEDBACK]
{previous_iteration_feedback_json}

[PLANNING CONTRACT]
{planning_contract_json}

[PREVIOUS STRATEGY MEMORY]
{previous_strategy_memory_json}
\end{lstlisting}

\begin{lstlisting}[style=stasboPrompt, caption={Strategy-planning response format. One alternative is shown.}, label={lst:app_planning_response}]
{
  "persistent_summary": {
    "reasoning": "<brief evidence assessment>",
    "evidence_summary": "<cumulative trajectory and unresolved comparisons>",
    "last_step_evaluation": "<intent, behavior, outcome, and implication>"
  },
  "behavioral_plan": {
    "reasoning": "<justification for retaining or revising intentions>",
    "overall_strategy": "<near-term strategy and revision evidence>",
    "current_focus": "<useful current comparison>",
    "alternatives": [
      {
        "plan_id": "<required ID>",
        "action": "<one-evaluation behavior>",
        "purpose": "<qualified benefit>",
        "candidate_preference": "<observable alignment cues>",
        "revision_condition": "<obtainable reassessment evidence>"
      }
    ]
  },
  "reasoning": "<concise additional tradeoffs>"
}
\end{lstlisting}

The trajectory summary and behavioral-plan fields form a complete replacement of the previous reusable strategy memory. The three \texttt{reasoning} fields contain concise justifications archived with the generation; they are removed from the strategy memory passed to subsequent calls. The alternatives are unordered intentions for one subsequent evaluation, rather than a priority list.

\paragraph{Configuration shortlisting.}
The shortlisting query supplies the updated strategy, fitted surrogate specifications, and a construction contract. The implementation uses the name \texttt{construction} for the operation termed configuration shortlisting in the paper; the corresponding query heading is retained below. The construction contract provides the required plan identifiers, the number of acquisitions per plan, the observed score range, and the admissible acquisition families with their parameter definitions and ranges.

\begin{lstlisting}[style=stasboPrompt, caption={Configuration-shortlisting query template.}, label={lst:app_shortlisting_query}]
[ENVIRONMENT]
{environment_json}

[CURRENT CONTEXT]
{decision_context_json}

[RECENT RECORDS]
{recent_records_json}

[PREVIOUS ITERATION FEEDBACK]
{previous_iteration_feedback_json}

[SURROGATE PROFILES]
{surrogate_profiles_json}

[CONSTRUCTION CONTRACT]
{construction_contract_json}

[STRATEGY MEMORY]
{strategy_memory_json}
\end{lstlisting}

\begin{lstlisting}[style=stasboPrompt, caption={Configuration-shortlisting response format. One plan and one acquisition are shown; the main setting requires three plans and three acquisitions per plan.}, label={lst:app_shortlisting_response}]
{
  "plan_configurations": [
    {
      "plan_id": "<required ID>",
      "surrogate_profile": "<available profile>",
      "surrogate_reasoning": "<qualified model assumption>",
      "acquisitions": [
        {
          "family": "ei",
          "parameters": {
            "xi_frac": 0.04
          },
          "reasoning": "<conditional tradeoff>"
        }
      ]
    }
  ],
  "reasoning": "<overall tradeoffs>"
}
\end{lstlisting}

For UCB, an acquisition entry instead uses \texttt{"family": "ucb"} and \texttt{"parameters": \{"kappa": 1.0\}}. EI accepts only \texttt{xi\_frac} in $[0, 1]$, and UCB accepts only \texttt{kappa} in $[0, 20]$. Identical family--parameter pairs cannot repeat within a plan, while overlap across plans is allowed. There is no prescribed allocation between EI and UCB. The returned rationales are archived but are not forwarded to the final selector. Validation and exact configuration deduplication produce $\widetilde{\mathcal C}_t$.

\paragraph{Joint configuration selection.}
The selection query presents the current strategy and calculated context together with exact-point candidate groups, their generating configurations, static surrogate definitions, and generation failures. The selection contract supplies the admissible current plan identifiers.

\begin{lstlisting}[style=stasboPrompt, caption={Joint configuration-selection query template.}, label={lst:app_selection_query}]
[ENVIRONMENT]
{environment_json}

[CURRENT CONTEXT]
{decision_context_json}

[RECENT RECORDS]
{recent_records_json}

[SURROGATE DEFINITIONS]
{surrogate_definitions_json}

[CANDIDATE DIAGNOSTICS]
{candidate_diagnostics_json}

[GENERATION FAILURES]
{generation_failures_json}

[SELECTION CONTRACT]
{selection_contract_json}

[STRATEGY MEMORY]
{strategy_memory_json}
\end{lstlisting}

\begin{lstlisting}[style=stasboPrompt, caption={Joint configuration-selection response format.}, label={lst:app_selection_response}]
{
  "configuration_id": "<successful configuration ID>",
  "selected_plan_id": "<allowed plan ID>",
  "decision_rationale": "<usefulness, alternative, and uncertainty>",
  "plan_check": {
    "requirements": [
      {
        "requirement": "<plan cue>",
        "status": "met",
        "evidence": "<supporting fact or uncertainty>"
      }
    ],
    "limitations": "<partial alignment and unknown objective>"
  }
}
\end{lstlisting}

The selected configuration must identify a successfully generated cached candidate. The LLM does not return new coordinates. The associated plan may be any current plan, independently of the configuration's originating plans, or JSON \texttt{null}. For an absent association, the prompt instructs the model to use \texttt{"selected\_plan\_id": null} and an empty \texttt{requirements} list. Requirement statuses are \texttt{met}, \texttt{not\_met}, or \texttt{uncertain}. The alignment assessment is advisory and does not establish objective improvement. The selection rationale and assessment are retained in the subsequent decision record.

\paragraph{Response validation and retries.}
Responses are checked for required fields, admissible identifiers, and valid parameter values. On an invalid response, the controller appends a \texttt{[VALIDATION FEEDBACK]} section containing \texttt{previous\_output\_error} to the query and requests a corrected response. The main setting permits at most five response attempts per logical decision, including the first attempt. The full system instructions and query templates will be provided with the supplementary implementation.

\subsection{Numerical BO Implementation}
\label{app:implementation_details:numerical_bo_implementation}

\paragraph{GP fitting.}
STaSBO and the fixed BO baselines use exact single-output GPs implemented with BoTorch and GPyTorch. Each surrogate combines its specified base kernel with a learned output scale and a Gaussian likelihood with learned noise. Inputs are normalized using the training-data coordinate ranges, and outputs are standardized using the observed mean and standard deviation, with safeguards for degenerate data. Models use double precision and are fitted by maximizing the exact marginal log-likelihood. At each STaSBO iteration, each available surrogate profile is fitted once to the current observations. Its fitted model is cached and reused across acquisition specifications.

\paragraph{Continuous acquisition optimization.}
STaSBO, BOs, the acquisition-portfolio methods, and BOOST's outer query generation use a shared continuous proposal procedure. LMABO and CAKE retain the acquisition-optimization routines of their respective reference implementations. The shared procedure samples 10,000 feasible points per iteration and scores the relevant acquisition on this pool. Box-domain sampling is uniform; simplex-domain sampling respects the composition constraint. The highest-scoring eligible points initialize up to twenty continuous-optimization restarts, with at most 500 optimizer iterations and restart batches of five. Optimization is performed in coordinates normalized to the declared search bounds. OER simplex constraints are imposed as linear equalities. The best valid refined point is compared with the best eligible sampled point, and the higher-scoring candidate is retained. If refinement fails, the sampled-pool candidate remains available.

\paragraph{Acquisition parameterization and candidate handling.}
For STaSBO's EI configurations, the improvement margin is
\[
    \xi_t = \xi_{\mathrm{frac}} \left( \max_{(\mathbf{x}, y) \in \mathcal D_t} y - \min_{(\mathbf{x}, y) \in \mathcal D_t} y \right).
\]
EI is evaluated through its numerically stable logarithmic form. UCB uses $\mu_t(\mathbf{x})+\kappa\sigma_t(\mathbf{x})$, corresponding to BoTorch's parameter $\beta=\kappa^2$. Candidates within $10^{-6}$ of an existing observation in unit-box maximum-coordinate distance are excluded. STaSBO caches each configuration's candidates before the final LLM decision. The selected cached point is evaluated once.

\subsection{Ablation Experiment Details}
\label{app:implementation_details:ablations}

\paragraph{Evaluation protocol.}
We compare full STaSBO with eight variants on Ackley-8D. Each condition uses ten seeds, indexed from 0 to 9, with the same random shift on the optimum location and five Latin-hypercube initial observations for each seed. Each run performs 45 additional objective evaluations. All variants use \texttt{gpt-6-luna} with medium reasoning effort and retain the problem description. We report raw final simple regret as the mean and sample standard deviation across seeds. The numerical BO settings follow Appendix~\ref{app:implementation_details:numerical_bo_implementation}. Table~\ref{tab:ablation_definitions} specifies the ablation modifications.

\begin{table}[h]
    \centering
    \small
    \caption{Definitions of the eight STaSBO ablations. All variants retain numerical BO candidate generation and evaluate one selected candidate per iteration.}
    \label{tab:ablation_definitions}
    \setlength{\tabcolsep}{5pt}
    \renewcommand{\arraystretch}{1.12}
    \begin{tabularx}{\linewidth}{@{}>{\raggedright\arraybackslash}p{0.28\linewidth}>{\raggedright\arraybackslash}X@{}}
        \toprule
        Variant & Modification relative to full STaSBO \\
        \midrule
        
        \multicolumn{2}{@{}l}{\textit{Joint Control}} \\
        \addlinespace[2pt]
        
        Acquisition-only control &
        Fix the surrogate profile to Mat\'ern-$5/2$ ARD while continuing to fit its GP hyperparameters and expose its fit evidence. Each plan nominates three acquisition specifications, including their parameter values. \\
        
        \addlinespace
        Surrogate-only control &
        Fix the acquisition to EI with $\xi_{\mathrm{frac}}=0$. Each plan nominates three distinct surrogate profiles, all paired with this fixed acquisition. Fit evidence and candidate diagnostics remain available. \\
        
        \midrule
        \multicolumn{2}{@{}l}{\textit{Planning and Memory}} \\
        \addlinespace[2pt]
        
        Planning from observations &
        Generate a new trajectory summary and behavioral plans from the complete observation history and numerical state at each iteration. Exclude previous strategy memory, decision records, and previous-iteration proposal feedback from LLM inputs. \\
        
        \addlinespace
        w/o trajectory summary $u_t$ &
        Neither generate nor supply the trajectory summary. Retain behavioral plans, including the previous plans used during planning, together with recent decision records and previous-iteration proposal feedback. \\
        
        \addlinespace
        w/o planning &
        Remove the planning call and exclude strategy memory, decision records, and previous-iteration proposal feedback from LLM inputs. Shortlisting and selection operate directly on observation history and numerical evidence, without behavioral plans or plan-alignment assessments. \\
        
        \midrule
        \multicolumn{2}{@{}l}{\textit{Numerical Evidence}} \\
        \addlinespace[2pt]
        
        w/o fit evidence &
        Remove current fitted numerical descriptors from the shortlisting input. Static surrogate specifications and fit availability remain visible. Numerical GP fitting and candidate generation are retained. \\
        
        \addlinespace
        w/o diagnostics (coords. retained) &
        Remove current candidates' normalized distances to the incumbent and nearest observation, together with their behavior labels. Candidate coordinates, exact-point grouping, and generating configurations remain visible. \\
        
        \addlinespace
        w/o diagnostics and coords. &
        Additionally remove current candidate coordinates, candidate-group identities, and grouping information. The selector receives a flat list of successful configurations, their acquisition specifications, and originating plan identifiers. \\
        
        \bottomrule
    \end{tabularx}
\end{table}

\paragraph{Joint configuration control.}
We ablate joint control by restricting adaptation to one BO component. Acquisition-only control fixes the surrogate profile to Mat\'ern-$5/2$ ARD. For each behavioral plan, the controller proposes three acquisition specifications, choosing EI with $\xi_{\mathrm{frac}}\in[0,1]$ or UCB with $\kappa\in[0,20]$. Surrogate-only control instead fixes the acquisition to EI with $\xi_{\mathrm{frac}}=0$ throughout the run. Each plan nominates three distinct surrogate profiles from the available seven-profile library, all paired with this acquisition. Profiles may be reused across plans, and duplicate configurations are merged, yielding at most seven distinct configurations in surrogate-only control. In both variants, GP length scales, observation-noise variance, and kernel output scale are fitted to the current observations, and fitted descriptors remain available during shortlisting. Both variants retain strategy planning, decision-aware memory, and all three LLM decision stages. Numerical BO generates the shortlisted configurations' candidates, and the final selector uses the usual candidate diagnostics to choose one for evaluation.

\paragraph{Planning and memory inputs.}
Planning from observations supplies the complete chronological observation history to the planning call. The resulting current summary and plans are then passed to shortlisting and selection, without carrying forward earlier controller memory or decision feedback. Without planning, the complete observation history is supplied directly to both shortlisting and selection. The variant without $u_t$ retains recurrent behavioral plans and decision feedback, but no trajectory summary is generated or supplied at any stage. All three planning and memory variants retain current surrogate fit evidence and prospective candidate diagnostics. All conditions use three logical LLM decision calls per iteration, except the variant without planning, which uses two. Observations and numerical audit records remain stored regardless of which information is exposed to the LLM.

\paragraph{Scope of evidence removal.}
The evidence ablations modify the information presented to the LLM while preserving numerical observations, surrogate fitting, and candidate generation. Both candidate-information ablations generate and cache prospective candidates before final selection. Without candidate coordinates, any reported plan-alignment status is restricted to \texttt{uncertain}, since the selector cannot inspect the current candidate realization. Historical evaluated-point geometry and previous-iteration proposal feedback remain available in these evidence ablations. The interventions therefore concern current fitted evidence or current prospective candidate information, while decision history and strategy memory are retained.

\paragraph{Configuration budgets.}
Full STaSBO, planning from observations, the variant without $u_t$, and the evidence ablations use three plans, each nominating one surrogate and three acquisition specifications. Acquisition-only control retains three acquisition nominations per plan. Surrogate-only control instead uses three surrogate nominations per plan. Without planning, shortlisting produces three configuration groups, each pairing one surrogate with three acquisition specifications. These groups have no associated behavioral plans, and acquisitions are paired only with their group's surrogate. All conditions therefore allow nine configuration nominations before duplicate removal. Exact duplicate configurations are merged before candidate generation, so the numbers of distinct configurations and candidate points can differ across conditions. Surrogate-only control admits at most seven distinct configurations from the seven-profile surrogate library. Every condition evaluates only one selected candidate per BO iteration.

\subsection{Configuration Selection Analysis}
\label{app:implementation_details:selection_patterns}

\paragraph{Selection records and aggregation.}
The analysis uses the final configurations selected by full STaSBO across all 22 benchmark tasks, ten seeds per task, and 45 BO iterations per run, yielding 9,900 selection records. For each non-overlapping five-iteration window, we compute selection proportions within each run, average over the ten seeds within each task, and then average equally across the 22 tasks. The ``Overall'' bars apply the same procedure to all 45 iterations. Surrogate profiles and acquisition families are summarized separately.

\paragraph{Acquisition-parameter categories.}
Selected acquisition settings are grouped using the family-specific boundaries in Table~\ref{tab:selection_parameter_categories}. EI uses the margin fraction $\xi_{\mathrm{frac}}$. UCB uses the uncertainty multiplier $\kappa$. These categories are used only for descriptive analysis. The parameter categories do not directly determine candidate displacement, novelty, or objective improvement: candidate behavior also depends on the fitted surrogate and accumulated observations. The temporal distributions describe aggregate preferences across tasks and seeds, rather than a common schedule followed by every run.

\begin{table}[h]
    \centering
    \small
    \caption{ Parameter categories used in the configuration-selection analysis. The labels describe acquisition settings, not measured candidate behavior. }
    \label{tab:selection_parameter_categories}
    \renewcommand{\arraystretch}{1.12}
    \begin{tabular}{@{}lll@{}}
        \toprule
        Category & EI & UCB \\
        \midrule
        Exploitation-oriented & $\xi_{\mathrm{frac}}\leq 0.05$ & $\kappa\leq 1$ \\
        Balanced & $0.05<\xi_{\mathrm{frac}}\leq 0.20$ & $1<\kappa\leq 3$ \\
        Exploration-oriented & $\xi_{\mathrm{frac}}>0.20$ & $\kappa>3$ \\
        \bottomrule
    \end{tabular}
\end{table}

\section{Additional Results}
\label{app:additional_results}

This section complements the rank-based results in the main manuscript with task-wise normalized final regret. These values show the magnitude of performance differences within each benchmark.

\subsection{Final Normalized Regret}
\label{app:final_simple_regret}

Tables~\ref{tab:main_results_synthetic_regret} and~\ref{tab:main_results_simulation_regret} report task-level normalized final simple regret, $q_{m,j}(T)$, using the Random normalization defined in Appendix~\ref{app:implementation_details:evaluation_protocol}. Each entry divides the method's mean final regret by Random's mean final regret on the same task. Random therefore has value one for every task. The final column averages the normalized values within the corresponding suite. Lower values indicate better performance.

\begin{table*}[h]
    \centering
    \caption{Mean normalized final regret on synthetic mathematical benchmarks. Lower is better.}
    \label{tab:main_results_synthetic_regret}
    
    \small
    \setlength{\tabcolsep}{4pt}
    \renewcommand{\arraystretch}{1.12}
    \begin{tabular*}{\linewidth}{@{\extracolsep{\fill}}l*{13}{c}@{}}
        \toprule
        \multirow{2}{*}{Method} &
        \multicolumn{3}{c}{Ackley} & \multicolumn{3}{c}{Levy} & \multicolumn{3}{c}{Rastrigin} & \multicolumn{3}{c}{Rosenbrock} &
        \multirow{2}{*}{\shortstack{Avg.\\norm.\\regret}} \\
        \cmidrule(lr){2-4}\cmidrule(lr){5-7}\cmidrule(lr){8-10}\cmidrule(lr){11-13}
        & 5D & 8D & 15D & 5D & 8D & 15D & 5D & 8D & 15D & 5D & 8D & 15D & \\
        \midrule
        Random                                & 1.00 & 1.00 & 1.00 & 1.00 & 1.00 & 1.00 & 1.00 & 1.00 & 1.00 & 1.00 & 1.00 & 1.00 & 1.000 \\
        BO-EI                                 & 0.40 & 0.74 & 0.93 & 0.53 & 0.31 & 0.56 & 0.74 & 0.69 & 0.79 & 0.22 & 0.18 & 0.38 & 0.539 \\
        BO-UCB                                & 0.18 & 0.72 & 0.99 & 0.64 & 0.50 & 0.85 & 0.82 & 0.85 & 1.02 & 0.51 & 0.44 & 0.92 & 0.702 \\
        \midrule
        GP-Hedge                              & 0.80 & 0.90 & 0.97 & 0.40 & 0.30 & 0.64 & 0.62 & 0.58 & 0.74 & 0.18 & 0.20 & 0.40 & 0.561 \\
        No-PASt-BO                            & 0.78 & 0.89 & 0.97 & 0.34 & 0.36 & 0.59 & 0.46 & 0.79 & 0.78 & 0.14 & 0.22 & 0.34 & 0.555 \\
        TuRBO                                 & 0.48 & 0.74 & 0.93 & 0.42 & 0.37 & 0.46 & 0.53 & 0.73 & 0.75 & 0.08 & 0.09 & 0.25 & 0.486 \\
        SETUP-BO                              & 0.79 & 0.85 & 0.96 & 0.24 & 0.44 & 0.59 & 0.57 & 0.79 & 0.73 & 0.13 & 0.15 & 0.39 & 0.552 \\
        BOOST                                 & 0.40 & 0.66 & 0.93 & 0.61 & 0.44 & 0.51 & 0.64 & 0.72 & 0.76 & 0.17 & 0.19 & 0.23 & 0.521 \\
        \midrule
        LLM                                   & 0.65 & 0.91 & 0.95 & 0.73 & 0.44 & 0.52 & 0.54 & 0.62 & 0.70 & 0.51 & 1.33 & 1.16 & 0.756 \\
        LMABO                                 & 0.40 & 0.85 & 0.97 & 0.30 & 0.19 & 0.54 & 0.79 & 0.74 & 0.84 & 0.30 & 0.32 & 0.28 & 0.543 \\
        CAKE                                  & 0.39 & 0.74 & 0.97 & 0.22 & 0.45 & 0.48 & 0.62 & 0.73 & 0.72 & 0.15 & 0.23 & 0.37 & 0.504 \\
        \midrule
        STaSBO                                & 0.32 & 0.42 & 0.87 & 0.28 & 0.27 & 0.49 & 0.64 & 0.60 & 0.65 & 0.08 & 0.13 & 0.11 & 0.404 \\
        \bottomrule
    \end{tabular*}
    
    \vspace{1pt}
\end{table*}

\begin{table*}[h]
    \centering
    \caption{Mean normalized final regret on scientific and engineering application benchmarks. Lower is better.}
    \label{tab:main_results_simulation_regret}
    
    \small
    \setlength{\tabcolsep}{4pt}
    \renewcommand{\arraystretch}{1.12}
    \begin{tabular*}{\linewidth}{@{\extracolsep{\fill}}l*{11}{c}@{}}
        \toprule
        \multirow{2}{*}{Method} & \multirow{2}{*}{AgNP} & \multirow{2}{*}{Concrete} & \multirow{2}{*}{\shortstack{Crossed\\Barrel}} & \multirow{2}{*}{HPLC} & \multirow{2}{*}{P3HT} & \multicolumn{4}{c}{OER Plate} & \multirow{2}{*}{\shortstack{Lunar\\Lander}} & \multirow{2}{*}{\shortstack{Avg.\\norm.\\regret}} \\ 
        \cmidrule(lr){7-10}
        & & & & & & 3496 & 3851 & 3860 & 4098 & & \\
        \midrule
        Random                                & 1.00 & 1.00 & 1.00 & 1.00 & 1.00 & 1.00 & 1.00 & 1.00 & 1.00 & 1.00 & 1.000 \\
        BO-EI                                 & 0.66 & 0.77 & 0.41 & 0.75 & 0.52 & 0.56 & 0.59 & 0.57 & 0.30 & 0.60 & 0.572 \\
        BO-UCB                                & 0.67 & 0.78 & 0.47 & 0.72 & 0.36 & 0.50 & 0.33 & 0.49 & 0.18 & 0.93 & 0.544 \\
        \midrule
        GP-Hedge                              & 0.62 & 0.81 & 0.51 & 0.88 & 0.62 & 0.57 & 0.52 & 0.61 & 0.25 & 0.66 & 0.606 \\
        No-PASt-BO                            & 0.62 & 0.79 & 0.50 & 0.80 & 0.54 & 0.33 & 0.69 & 0.62 & 0.36 & 0.68 & 0.594 \\
        TuRBO                                 & 0.26 & 0.81 & 0.57 & 0.63 & 0.79 & 0.47 & 0.66 & 0.63 & 0.23 & 0.68 & 0.573 \\
        SETUP-BO                              & 0.54 & 0.79 & 0.57 & 0.84 & 0.59 & 0.52 & 0.68 & 0.66 & 0.35 & 0.73 & 0.626 \\
        BOOST                                 & 0.71 & 0.85 & 0.67 & 0.83 & 0.45 & 0.85 & 0.46 & 0.82 & 0.34 & 0.84 & 0.682 \\
        \midrule
        LLM                                   & 1.16 & 0.98 & 0.96 & 1.14 & 1.07 & 1.11 & 1.02 & 1.01 & 1.04 & 1.13 & 1.060 \\
        LMABO                                 & 0.58 & 0.82 & 0.41 & 0.71 & 0.58 & 0.39 & 0.47 & 0.63 & 0.29 & 0.43 & 0.530 \\
        CAKE                                  & 0.56 & 0.85 & 0.39 & 0.76 & 0.54 & 0.34 & 0.61 & 0.73 & 0.28 & 0.61 & 0.568 \\
        \midrule
        STaSBO                                & 0.45 & 0.78 & 0.39 & 0.71 & 0.39 & 0.35 & 0.52 & 0.61 & 0.18 & 0.64 & 0.502 \\
        \bottomrule
    \end{tabular*}                    
    
    \vspace{1pt}
\end{table*}


\end{document}